\documentclass[runningheads]{llncs}

\usepackage{eccv}

\usepackage{eccvabbrv}

\usepackage{graphicx}
\usepackage{booktabs}

\usepackage[accsupp]{axessibility}  

\usepackage{url}            
\usepackage{booktabs, colortbl}       
\usepackage{amsfonts}       
\usepackage{nicefrac}       
\usepackage{microtype}      
\usepackage[inline]{enumitem}

\usepackage{amsmath}

\usepackage{wrapfig, lipsum, booktabs}

\usepackage{algorithmicx}
\usepackage{algorithm}
\usepackage[noend]{algpseudocode}
\usepackage{mathtools} 

\usepackage{float}
\usepackage{soul}
\usepackage{multirow}
\usepackage{IEEEtrantools}

\usepackage{booktabs}
\usepackage{bbm}
\usepackage{mathrsfs}
\usepackage{graphicx}
\usepackage[export]{adjustbox}
\usepackage{blindtext}

\usepackage{wrapfig}

\usepackage{tabulary}
\usepackage{tabularx}

\usepackage{subcaption}

\newcommand{\supp}{\textcolor{black}{supplementary}}

\newcommand{\newpara}[1]{\noindent \textbf{#1} \hspace{2pt} }
\newcommand{\deit}{DeiT}
\newcommand{\graycolor}[1]{\textcolor{gray}{#1}}

\newenvironment{tight_enumerate}{
\begin{enumerate}[leftmargin=15pt]
  \setlength{\topsep}{0pt}
  \setlength{\itemsep}{0pt}
  \setlength{\parskip}{0pt}
  \setlength{\parsep}{0pt}
}{\end{enumerate}}

\usepackage{url}

\usepackage{hyperref}

\usepackage[capitalize]{cleveref}
\crefname{section}{Sec.}{Sec.}
\crefname{table}{Tab.}{Tab.}

\begin{document}

\title{UDA-Bench: Revisiting Common Assumptions in \\ Unsupervised Domain Adaptation Using a Standardized Framework}

\titlerunning{UDA-Bench}

\author{Tarun Kalluri \and
Sreyas Ravichandran \and
Manmohan Chandraker}

\authorrunning{T.~Kalluri et al.}

\institute{UC San Diego \\
\url{https://github.com/ViLab-UCSD/UDABench_ECCV2024}}

\maketitle

\begin{abstract}
    In this work, we take a deeper look into the diverse factors that influence the efficacy of modern unsupervised domain adaptation (UDA) methods using a large-scale, controlled empirical study. To facilitate our analysis, we first develop UDA-Bench, a novel PyTorch framework that standardizes training and evaluation for domain adaptation enabling fair comparisons across several UDA methods. Using UDA-Bench, our comprehensive empirical study into the impact of backbone architectures, unlabeled data quantity, and pre-training datasets reveals that: (i) the benefits of adaptation methods diminish with advanced backbones, (ii) current methods underutilize unlabeled data, and (iii) pre-training data significantly affects downstream adaptation in both supervised and self-supervised settings. In the context of unsupervised adaptation, these observations uncover several novel and surprising properties, while scientifically validating several others that were often considered empirical heuristics or practitioner intuitions in the absence of a standardized training and evaluation framework. The UDA-Bench framework and trained models are \href{https://github.com/ViLab-UCSD/UDABench_ECCV2024}{publicly available}.
\end{abstract}

\section{Introduction}

Deep neural networks for image classification often suffer from dataset bias where accuracy significantly drops if the test-time data distribution does not match that of training, which often happens in real-world applications. To overcome the infeasibility of collecting labeled data from each application domain, a suite of methods have been recently proposed under the umbrella of unsupervised domain adaptation (UDA)~\cite{hoffman2013efficient, long2015learning, long2017deep, bousmalis2017unsupervised, bousmalis2016domain, CDAN, DANN, saito2017adversarial, saito2018maximum, zhang2019bridging, hoffman2018cycada, xu2019larger, jin2020minimum, sharma2021instance, kang2019contrastive, wei2021toalign, kalluri2022memsac, kalluri2024tell, berthelot2021adamatch, zhu2023patchmix} that allow training using only unlabeled data from the target domain of interest while leveraging supervision from a different source domain with abundant labels. 
%
%
\begin{figure}[htbp]
    \centering
    \begin{minipage}[b]{.14\textwidth}
        \centering
        \includegraphics[width=\linewidth]{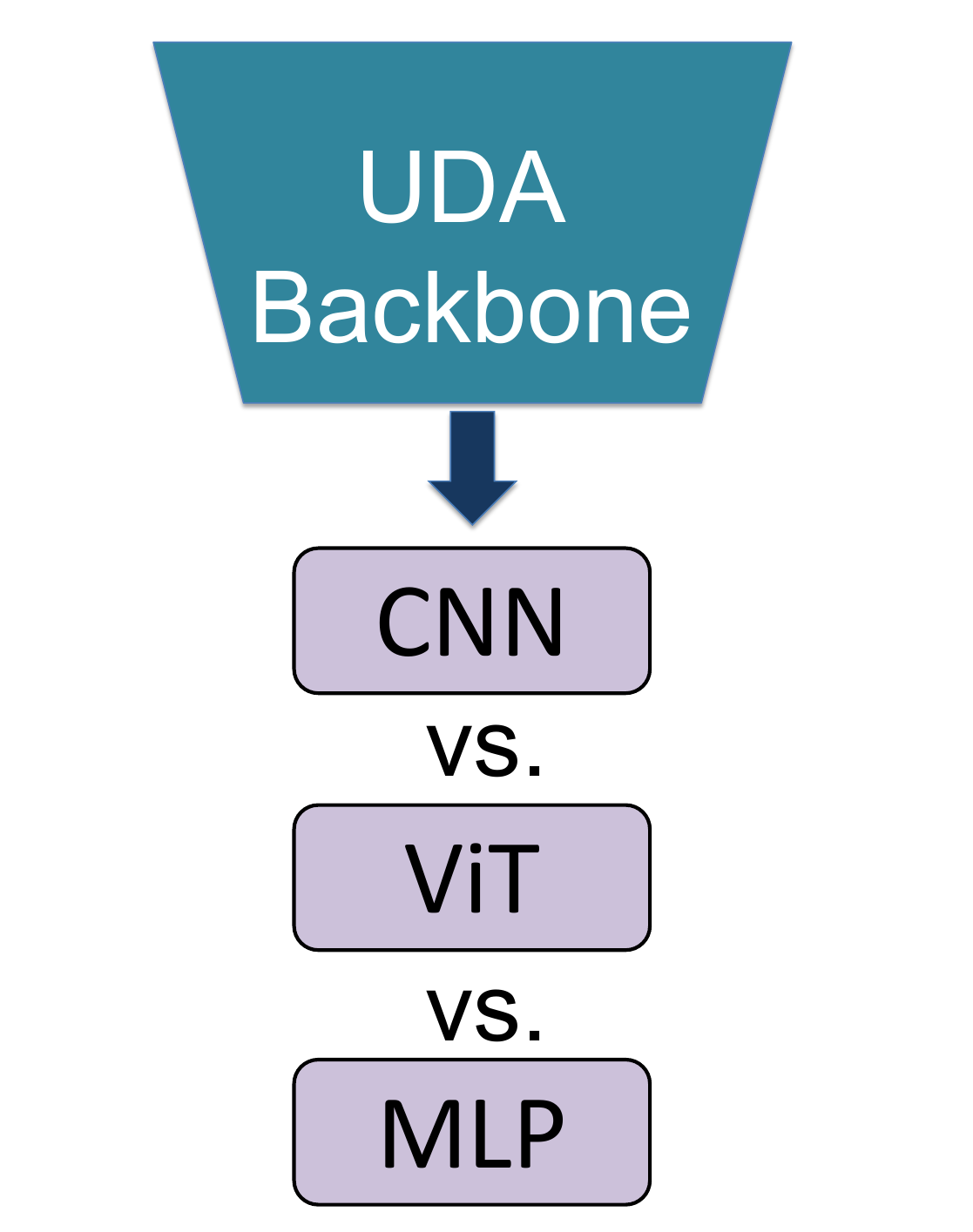}
        \subcaption{{Backbone}}
        \label{fig:intro_backbone}
    \end{minipage}
    \hfill
    \begin{minipage}[b]{.4\textwidth}
        \centering
        \includegraphics[width=\linewidth]{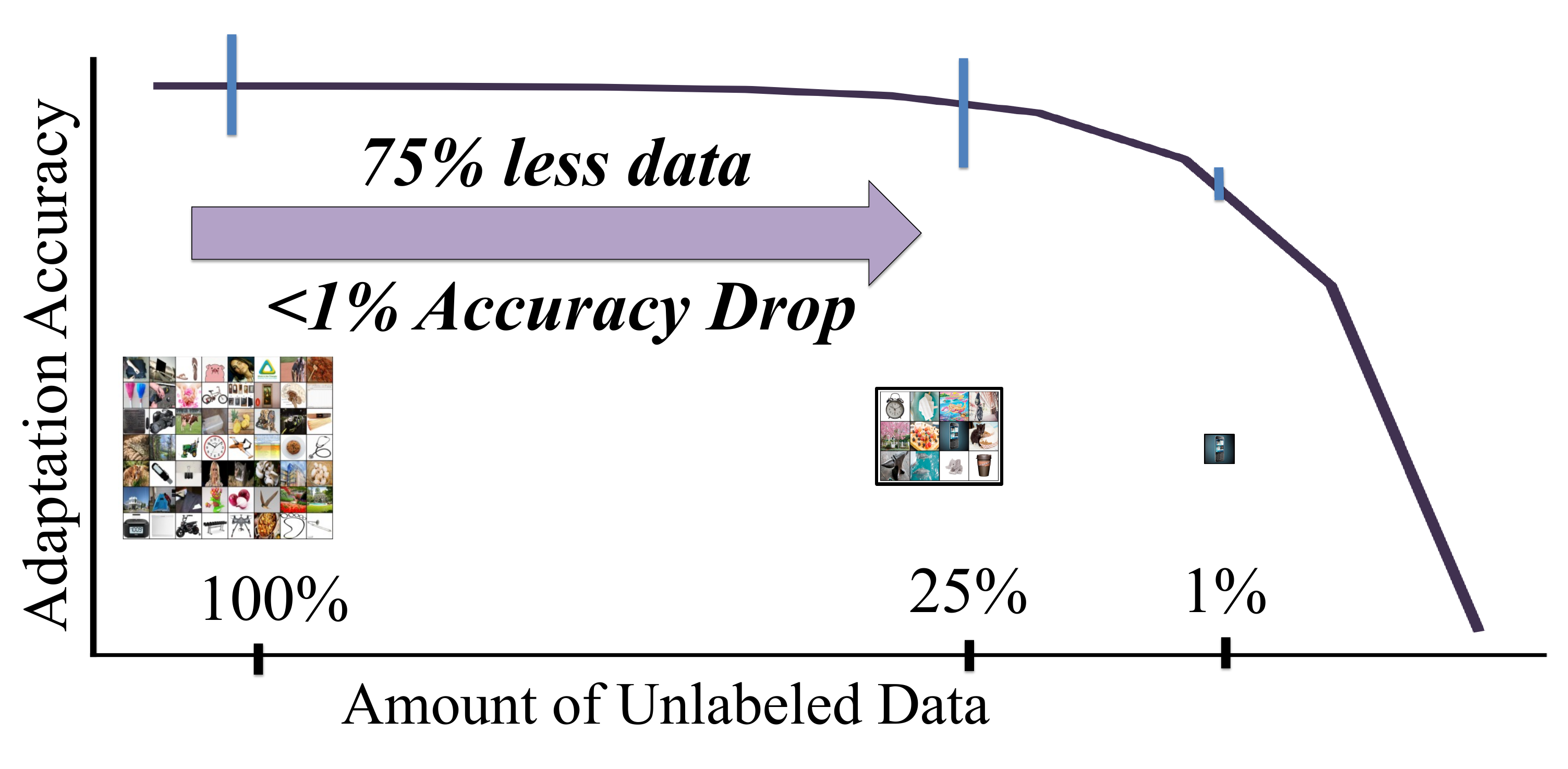}
        \subcaption{{Sample Efficiency of Target Data}}
        \label{fig:intro_vol}
    \end{minipage}
    \hfill
    \begin{minipage}[b]{.4\textwidth}
        \centering
        \includegraphics[width=\linewidth]{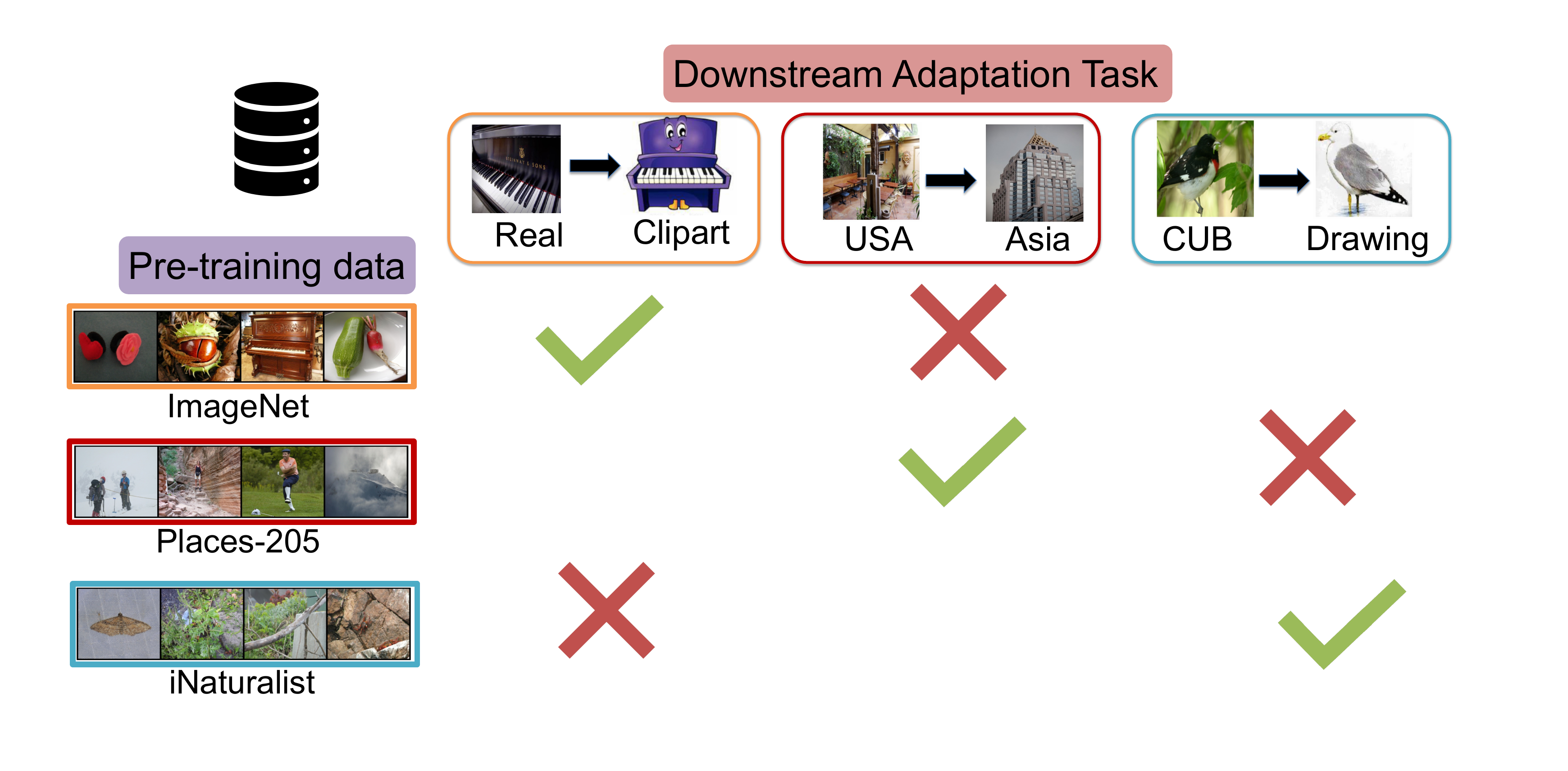}
        \subcaption{{Type of Pre-Training Data}}
        \label{fig:intro_pre}
    \end{minipage}
    \caption{\textbf{A summary of our contributions.} We examine the effectiveness of SOTA UDA approaches using our proposed framework UDA-Bench by revisiting the role of backbone architectures (\cref{fig:intro_backbone}, \cref{sec:backbone}), unlabeled data (\cref{fig:intro_vol}, \cref{sec:unlabeled_data}) and pre-training data (\cref{fig:intro_pre}, \cref{sec:pretraining}) with several useful observations.}
    \label{fig:result_summary}
    \vspace{-19pt}
\end{figure}
These UDA methods have been greatly successful in improving the target accuracy on benchmark datasets under a variety of distribution shifts~\cite{saenko2010adapting, peng2017visda, venkateswara2017deep, caputo2014imageclef, peng2019moment}. While literature in the area has predominantly focused on proposing new algorithms or loss functions, a holistic understanding of several fundamental assumptions that influence real-world effectiveness of domain adaptation has been lacking. In this paper, we address this through a large-scale empirical study of three major factors that potentially influence performance the most, namely,
\begin{enumerate*}

    \item {\bf Choice of backbone architecture:} 
    With recent advances in architecture designs such as vision transformers~\cite{dosovitskiy2020image, touvron2021training, liu2021swin} and improved CNNs~\cite{liu2022convnet} we study which architectures suit domain transfer, and verify compatibility of existing adaptation methods with these backbones.

    \item {\bf Amount of unlabeled data:} 
    Since the promise of unsupervised adaptation rests on its potential to leverage unlabeled target domain data, we study how much unlabeled data can really be digested by the adaptation methods.

    \item {\bf Nature of pre-training data:}
    We examine whether pre-training the backbone on similar data as the downstream adaptation task is more beneficial than commonly adopted ImageNet pre-training across several supervised and self-supervised pre-training strategies.
\end{enumerate*}

We believe that such insights into the behavior of UDA methods have been previously hindered due to varying choices of adaptation-independent factors like initialization, learning algorithm and batch sizes. To address this, we first propose UDA-Bench, a new PyTorch framework that standardizes these factors across multiple UDA methods and offers a unified training and evaluation platform for unsupervised adaptation. Using this framework, we study various UDA methods for image classification under different factors of variation. 
{Among prior works which shared similar motivations as ours~\cite{kim2022broad}, the absence of standardized evaluation limits fair comparisons between UDA methods, where our distinction lies in establishing such a framework for consistent UDA training and evaluation. Through our analysis, we discover several new insights, while scientifically validating several phenomenon which were only considered empirical heuristics or practitioner intuitions due to the lack of a standardized approach. }
These are outlined in \cref{fig:result_summary}, and can be summarized as follows:
 \begin{tight_enumerate}

    \item Recent advancements in vision transformers such as Swin~\cite{liu2022swin} and DeiT~\cite{touvron2022deitiii} exhibit superior robustness against diverse domain shifts when compared to the conventional choice of ResNet-50 (see \cref{tab:arch_domain_robustness}). However, incorporating these advancements into current UDA methods tends to diminish their benefits, leading to significant changes to the relative ranking among the methods. As a result, \textit{older and simpler UDA methods often achieve comparable or even superior accuracies compared to more recent methods} (see \cref{fig:arch_da_ablation} and \cref{sec:backbone}).
    
    \item Reducing the amount of unlabeled target data by up to 75\% resulted in only a 1\% decrease in target accuracy across all UDA methods studied (see \cref{fig:datavol_ablation}), suggesting that \textit{that current UDA methods saturate quickly, and are not well-equipped to exploit the increasing availability of inexpensive unlabeled data} (see \cref{sec:unlabeled_data}). This observation also contradicts the prevailing theory underpinning modern UDA research proposed in Ben-David et. al.~\cite{ben2010theory}, which suggests an inverse relation between the amount of unlabeled target data and target error, highlighting the discrepency between theory and practice.

    \item Pre-training data matters for downstream adaptation, but in different ways for supervised and self-supervised pre-training. In supervised setting, \textit{pre-training on similar data as the downstream adaptation task significantly improves the accuracy} compared to standard ImageNet pre-training (see \cref{tab:pt_sup}).  

    \item In self-supervised setting, \textit{object-centric pre-training datasets enhance accuracy for object-centric adaptation}, while scene-centric pre-training datasets are better suited for scene-centric tasks (see \cref{tab:pt_ssl}). This trend holds across different types of pre-text tasks in self-supervised pre-training (see \cref{sec:pretraining}).

\end{tight_enumerate}

Through a comprehensive analysis using our unified training and evaluation framework, our recommendations serve a dual purpose - enabling researchers in identifying future opportunities for developing more effective adaptation algorithms with fair comparisons, as well as guiding practitioners in maximizing the benefits derived from current UDA methods. Our framework is \href{https://github.com/ViLab-UCSD/UDABench_ECCV2024}{publicly available} to continue improving our understanding of UDA methods.

\section{Related Works}

\newpara{Unsupervised Domain Adaptation} 
A majority of works in unsupervised adaptation aim to minimize some notion of divergence between the source and target domains estimated using unlabeled samples~\cite{ben2010theory, ben2006analysis}. Prior works studied various divergence metrics like MMD distance~\cite{long2015learning, long2017deep, hsu2015unsupervised, yan2017mind, baktashmotlagh2016distribution, pan2010domain, long2013transfer, tzeng2014deep}, higher-order correlations~\cite{morerio2017minimal, sun2016deep, sun2016return, jin2020minimum} or optimal-transport~\cite{courty2017optimal, redko2019optimal, damodaran2018deepjdot}, but adversarial discriminative approaches~\cite{DANN, tzeng2014deep, tzeng2017adversarial, xie2018learning, CDAN, tzeng2015simultaneous, chen2019progressive, saito2018maximum} have been the most popular. 
More recent works address the issue of noisy alignment with global domain discrimination~\cite{kumar2018co} using category-level~\cite{saito2017asymmetric, pei2018multi, kang2019contrastive, na2021fixbi, du2021cross, wei2021toalign, cui2020hda, prabhu2021sentry}, instance-level~\cite{sharma2021instance, kalluri2022memsac}, consistency-based~\cite{berthelot2021adamatch}, language-guided~\cite{kalluri2024tell} or cross-attention~\cite{xu2021cdtrans, zhu2023patchmix} based techniques. The primary focus of most of these works is on algorithmic innovations to improve adaptation. Instead, our emphasis in this paper lies in identifying several key method-agnostic factors that impact performance of UDA methods, and conducting a comprehensive empirical study along these factors for a better understanding of these methods. While domain adaptive semantic segmentation is also popular~\cite{tsai2018learning, hoffman2018cycada, vu2019advent, kalluri2022cluster, Lai2022DecoupleNetDN}, we restrict focus on adaptation methods for image-classification in this paper.

\newpara{Comparative Studies and Benchmarks} Many recent works aim to enhance our understanding of the factors impacting the success of state-of-the-art methods through carefully crafted empirical analysis. A common theme in these works is to keep the algorithm itself fixed, but study several other factors which hold non-trivial importance in determining the performance of the algorithm. Within computer vision, these works span the areas of semi-supervised learning~\cite{oliver2018realistic}, SLAM~\cite{nardi2014introducing}, metric learning~\cite{musgrave2020metric, roth2020revisiting}, transfer learning~\cite{mensink2021factors}, domain generalization~\cite{gulrajani2020search}, optimization algorithms~\cite{choi2019empirical}, few-shot learning~\cite{chen2019closer}, contrastive learning~\cite{cole2022does}, GANs~\cite{lucic2018gans}, fairness~\cite{goyal2022vision} and self-supervised learning~\cite{goyal2019scaling, newell2020useful, goldblum2023battle}. Prior works also established standardized benchmarks to facilitate fair comparisons and quick prototyping~\cite{musgrave2020metric, tancik2023nerfstudio, nardi2014introducing}.
Our work follows suit, where we develop a unified framework for UDA methods, and devise a controlled empirical study to revisit several standard training choices in unsupervised adaptation.

The works closest to ours in domain adaptation are \cite{kim2022broad}, which carries UDA study but without a unified training framework, \cite{musgrave2022benchmarking, musgrave2021unsupervised}, which study UDA methods through fair validation methods and \cite{kareer2024using} which studies adaptation for video segmentation. Different from these, our work lays emphasis on several other key factors that impact adaptation such as architectures, quantity of unlabeled data and nature of pretext data used in pre-training through design of a new standardized evaluation framework.

\section{Analysis Setup}
\label{sec:analysis_setup}

The task of unsupervised domain adaptation (UDA) aims to improve performance on a certain target domain with only unlabeled samples ($D_t{=}\{X_t\}$) by leveraging supervision from a different labeled source domain $D_s{=}\{X_s, y_s\}$.
We assume that the source images are drawn from $X_s {\sim} P_s$, and target images from $X_t {\sim} P_t$.
We assume a covariate shift~\cite{ben2010theory} between the domains, which arises when $P_s {\neq} P_t$, although other forms of shift have also been studied in literature~\cite{tan2020class, DBLP:journals/pami/Azizzadenesheli22, garg2020unified, alabdulmohsin2023adapting}. The task of UDA is then to learn a predictive model using $\{X_s,X_t,y_s\}$ to improve performance on test samples from the target domain $P_t$.
%
While recent literature focuses on novel training algorithms or loss functions to improve transfer, this paper aims to study their effectiveness under several important but often overlooked axes of variations pertaining to backbone architectures, unlabeled data quantity and backbone pre-training strategies. 

\begin{figure*}[t]
\begin{center}
    \begin{minipage}[b]{0.75\textwidth}
        \centering
        \includegraphics[width=\textwidth]{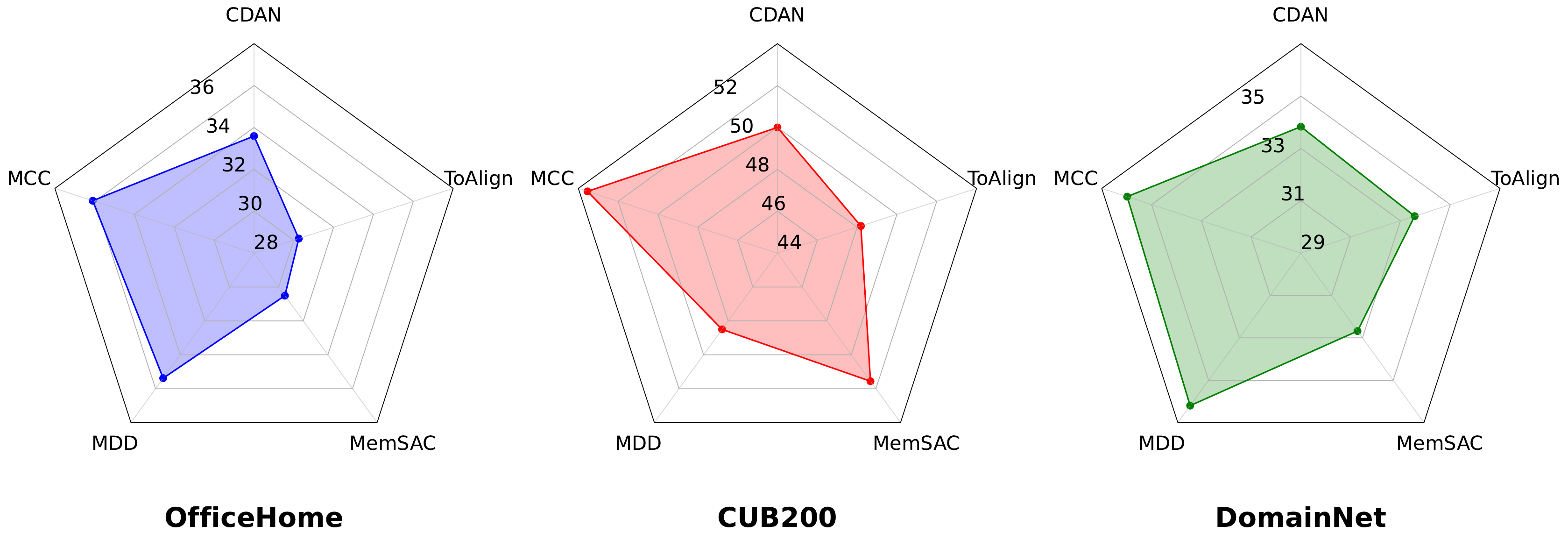}
        \vspace{-8pt}
    \end{minipage}
    \captionsetup{width=\textwidth, font=footnotesize}
    \caption{{\bf Need for UDA-Bench.} We illustrate the disparity between various codebases proposed for prior UDA methods by highlighting the different accuracy numbers obtained for a plain source only model. Computed without any adaptation, it should ideally match across implementations which is clearly not the case. To enable fair comparisons across UDA methods, we propose UDA-Bench, a new PyTorch framework to standardize training and evaluation across various methods.} 
    \label{fig:joint_plain_acc}
    \vspace{-2em}
\end{center}
\end{figure*}

\newpara{The Need for UDA-Bench Framework} 
Ensuring fair comparisons between different UDA methods necessitates controlling algorithm-independent factors during training and inference. 
However, we identify a problematic practice in most UDA methods where they are trained on different frameworks with different choices in various training hyper-parameters and settings, making fair comparison across these works difficult. 
To highlight this issue, we compute the plain source-only accuracy using original code-bases of various UDA algorithms in \cref{fig:joint_plain_acc} (the links to the open-source code for each of these methods are given in the \supp{}). Essentially, we take the open-source code base for the methods, switch off all the adaptation losses, and train the model only on the source dataset to compute the target accuracy. Ideally, this accuracy, which acts as the baseline, should be the same across all the methods since it is independent of any adaptation. In practice, however, we observe that this baseline accuracy varies significantly between various UDA codebases, pointing to an underlying discrepency in various training choices adopted by these works unrelated to the adaptation algorithm itself. For example, unique to the respective methods, MDD~\cite{zhang2019bridging} uses a deeper MLP as a classifier, MCC~\cite{jin2020minimum} uses batchnorm layers in the bottleneck layer, CDAN~\cite{CDAN} uses 10-crop evaluation and AdaMatch~\cite{berthelot2021adamatch} uses stronger augmentation on source data.

To alleviate this issue, we create a new framework in PyTorch~\cite{paszke2017automatic} for domain adaptation called \textit{UDA-Bench} and implement several existing methods in this framework. Our framework standardizes different UDA methods with respect to adaptation-independent factors such as learning algorithm, network initialization and batch sizes while simultaneously allowing flexibility for incorporating algorithm-specific hyperparameters like loss coefficients and custom data loaders within a unified framework.
All our comparisons and analyses in this paper are implemented using this framework, while using the same adaptation-specific hyperparameters proposed in the original papers in our re-implementation. 
{We also verified that our re-implementations reproduced the original accuracies when using the hyper-parameters from the respective codebases.}
UDA-Bench, along with all our implementations, is \href{https://github.com/ViLab-UCSD/UDABench_ECCV2024}{publicly released} to the research community to enable fair comparisons and fast prototyping of UDA methods in future works.

\newpara{Axes of Variation} { We choose backbone architecture (\cref{sec:backbone}), amount of unlabeled data in the target (\cref{sec:unlabeled_data}) and the nature of data/algorithm used in pre-training the backbone (\cref{sec:pretraining}) as the different axes of variation in our study. The deliberate focus on backbone, data size, and pre-training factors is driven by the recognition that these factors hold the most potential to influence deep learning training in general and UDA algorithms in particular, while also being the most understudied in prior UDA literature. By analyzing these factors, we seek to offer insights into salient properties of UDA and provide practical guidance for enhancing accuracy through optimal design choices.}

\newpara{Adaptation Methods} 
The selection of methods in our comparative study is not intended to be exhaustive of all the adaptation methods proposed in the literature thus far. Instead, we aim to provide a representative sample of works spanning a diverse range of model families from standard to state-of-the-art, although our inferences should readily transfer to any UDA method. In particular, the types of UDA methods we study include \textit{adversarial} (DANN~\cite{DANN}, CDAN~\cite{CDAN}), \textit{non-adversarial} (MDD~\cite{zhang2019bridging}, MCC~\cite{jin2020minimum}, DALN~\cite{Chen_2022_CVPR}), \textit{consistency-based} (MemSAC~\cite{kalluri2022memsac}, AdaMatch~\cite{berthelot2021adamatch}), \textit{alignment-based} (ToAlign~\cite{wei2021toalign}) and \textit{pseudo-label based}~\cite{zhu2023patchmix} methods. 
In the \supp{}, we show that the inferences made in our study also extend to several other adaptation methods (such as BSP~\cite{chen2019transferability}, ILADA~\cite{sharma2021instance}, AFN~\cite{xu2019larger} and MCD~\cite{saito2018maximum}).

\newpara{Adaptation Datasets} Following popular choices in UDA literature, we use visDA~\cite{peng2017visda}, OfficeHome~\cite{venkateswara2017deep}, DomainNet~\cite{peng2019moment} and CUB200~\cite{PAN} datasets in our analysis. VisDA studies synthetic to real transfer from 12 categories, OfficeHome contains 65 categories across four domains, DomainNet contains images from 345 categories from 6 domains while CUB200 is designed for fine-grained adaptation. In the \supp{}, we also show results using adaptation on TinyImageNet~\cite{le2015tiny} and variants~\cite{hendrycks2019benchmarking}. 

\newpara{Evaluation Metrics} We report results using the accuracy on the test set of the target domain while correcting for a problematic practice in prior literature. In most prior works using OfficeHome and CUB200 datasets, the same set of data doubles up as the unlabeled target used in training as well as the target test set used to report the results. To avoid possible over-fitting to target unlabeled data, we create separate train and test sets for these datasets (using a 90\%-10\% ratio), and use images from train set as labeled or unlabeled data during training and report final numbers on the unused test images. While this could lead slightly different numbers from those reported in the original papers, it also leads to fair comparison with the source-only baseline. 

\newpara{Hyper-parameters} In all our re-implementations of prior works, we use the default hyperparameters suggested by the original methods to keep the number of experiments manageable. Each method in the unlabeled data volume study (\cref{sec:unlabeled_data}) takes about 24 hours to run on an NVIDIA A10 GPU, so 8 methods, across 4 settings, 6 data fractions and 3 random trials costs $\sim$14000 GPU hours. Likewise, the experiments in \cref{sec:backbone} cost 18640 GPU hours and \cref{sec:pretraining} cost about 17356 GPU hours (including the pre-training). Incorporating experiments to seek optimal hyperparameters for several UDA methods on top of this would have incurred impractical levels of expenses. 



\section{Methodology and Evaluation}
\label{sec:methodology}

\begin{table*}[!tb]
    \centering
    \captionsetup{width=\textwidth, font=footnotesize}
    \caption{\textbf{Comparison of domain robustness of various vision architectures} on standard adaptation datasets. We use the source accuracy ($\lambda_s$) and the target accuracy ($\lambda_t$) of a model trained only on source data to calculate the relative drop in accuracy ($\sigma_{st}{=}100*(\lambda_s-\lambda_t)/\lambda_s$, lower the better). Swin transformer shows consistently better robustness to domain shifts on several benchmarks.}
    \label{tab:arch_domain_robustness}
    \begin{minipage}{0.91\textwidth}
    \resizebox{\textwidth}{!}{%
    \begin{tabular}{@{}cccccccccccccc@{}}
        \toprule
        Model & ResNet-50 & Swin-V2-t & ConvNext-t & ResMLP-s & \deit3-s && ResNet-50 & Swin-V2-t & ConvNext-t & ResMLP-s & \deit3-s \\
        \graycolor{\#Params} & \graycolor{24.12 M} & \graycolor{27.86 M} & \graycolor{28.10 M} & \graycolor{29.82 M} & \graycolor{21.86 M} && \graycolor{24.12 M} & \graycolor{27.86 M} & \graycolor{28.10 M} & \graycolor{29.82 M} & \graycolor{21.86 M} \\
        \cline{2-6} \cline{8-12}
         & \multicolumn{5}{c}{\underline{DomainNet (R$\rightarrow$C)}} && \multicolumn{5}{c}{\underline{CUB200 (CUB$\rightarrow$Draw)}} \\
        \rule{0pt}{3ex}  Source Accuracy ($\lambda_s, \uparrow$) & 81.86 & 85.99 & 84.37 & 82.68 & 84.52 && 81.00 &  87.75 & 85.88 & 84.62 & 88.12 \\
        Target Accuracy ($\lambda_t, \uparrow$) & 44.85 & 55.51 & 50.80 & 46.62 & 50.75 && 52.60 &  58.90 & 52.74 & 53.41 & 56.36 \\
        Relative Drop ($\sigma_{st}$, $\downarrow$) & 45.21 & \textbf{35.45} & \underline{39.78} & 43.61 & 39.95 && \underline{35.0} &  \textbf{32.88} & 38.50 & 36.88 & 36.05 \\
        Abs. Drop ($\lambda_s-\lambda_t$, $\downarrow$) & 37.01 & 30.48 & 33.57 & 36.06 & 33.77 && 28.40 & 28.85 & 33.14 & 31.21 & 31.76 \\
        \cline{2-6} \cline{8-12}
         & \multicolumn{5}{c}{\underline{OfficeHome (Ar$\rightarrow$Pr)}} && \multicolumn{5}{c}{\underline{GeoPlaces (USA$\rightarrow$Asia)}} \\
        \rule{0pt}{3ex} Source Accuracy ($\lambda_s, \uparrow$) & 60.10 & 76.17 & 74.72 & 69.69 & 71.76 && 57.17 & 63.11 & 60.39 & 58.99 & 61.65 \\
        Target Accuracy ($\lambda_t, \uparrow$) & 53.33 & 72.56 & 70.77 & 65.90 & 67.18 && 36.12 & 42.53 & 40.30 & 38.11 & 40.34 \\
        Relative Drop ($\sigma_{st}$, $\downarrow$) & 11.26 & \textbf{4.74} & \underline{5.29} & 5.44 & 6.38 && 36.82 & \textbf{32.61} & \underline{33.27} & 35.40 & 34.57 \\
        Abs. Drop ($\lambda_s-\lambda_t$, $\downarrow$) & 6.77 & 3.61 & 3.95 & 3.79 & 4.58 && 21.05 & 20.58 & 20.09 & 20.88 & 21.31 \\
        \bottomrule
    \end{tabular}
    }
    \end{minipage}%
    \vspace{-16pt}
\end{table*}

\subsection{Which backbone architectures suit UDA best?}
\label{sec:backbone}

\newpara{Motivation} 
%
Although ResNet-50~\cite{he2015deep} backbone is a widely adopted standard in domain adaptation research~\cite{CDAN, saito2018maximum, saito2017asymmetric, berthelot2021adamatch, kalluri2022memsac, wei2021toalign}, several recent architectures~\cite{dosovitskiy2020image, touvron2021training, liu2022convnet} have emerged as feasible alternatives with better performance. 
While a more recent method PMTrans~\cite{zhu2023patchmix} adopts a ViT backbone, all the prior methods were still compared using a ResNet-50 backbone. 
Therefore, we aim to study if the recent advances in vision transformers confer additional benefits to cross-domain transfer, and how ViT-specific methods~\cite{zhu2023patchmix} compare to classical methods while using a same backbone. While robustness properties of vision transformers to adversarial and out-of-context examples have been widely studied~\cite{bai2021transformers, bhojanapalli2021understanding, Shao2021OnTA, zhao2021robin, zhou2022understanding}, our analysis differs from these by focusing on the \textit{cross-domain robustness} properties of these architectures on standard UDA datasets and investigating their potential as an improved backbone for UDA methods.

\newpara{Experimental Setup} Along with ResNet-50, we choose four different vision architectures which showed great success on standard ImageNet classification benchmarks: DeiT~\cite{touvron2021training}, Swin~\cite{liu2021swin}, ResMLP~\cite{touvron2022resmlp}, and ConvNext~\cite{liu2022convnet}. We use newer versions of DeiT (DeiT-III~\cite{touvron2021training}) and Swin (Swin-V2~\cite{liu2021swin}) as they have better accuracy on ImageNet. We use the variants of these architectures which roughly have comparable number of parameters as ResNet-50, namely {DeiT-small}, {Swin-tiny}, {ResMLP-small} and {ConvNext-tiny}. All of them are pre-trained on ImageNet-1k, so their differences only arise from specific architectures. We use all pre-trained checkpoints from the timm library~\cite{rw2019timm} and architecture-specific training details are provided in the \supp{}. 



\vspace{4pt}
\newpara{Newer Architectures Show Better Domain Transfer}
For a model trained only on source-domain data (no adaptation), we use the accuracy on the source test-set ($\lambda_s$) and the accuracy on the target test-set ($\lambda_t$), to define relative cross-domain accuracy drop $\sigma_{st} {=} \tfrac{\lambda_s-\lambda_t}{\lambda_s}*100$. While this metric is sensitive to the absolute value of the source accuracy ($\lambda_s$), we nevertheless find that it serves as a good indicator of cross-domain robustness. Additionally, we also show the absolute accuracy drop from source to target ($\lambda_s{-}\lambda_t$) to discount the effect of original source accuracy. 
From \cref{tab:arch_domain_robustness}, vision transformer architectures have the least value of $\sigma_{st}$ (least cross-domain drops) indicating better robustness properties compared to CNNs or MLPs. Specifically, {Swin-V2-t pre-trained on ImageNet-1k showed least relative drop} ($\sigma_{st}$) across all the datasets.
Notably, on Real$\rightarrow$Clipart from DomainNet, using Swin backbone with plain source-only training alone yields 55.5\% accuracy, which is already higher than SOTA UDA methods that use ResNet-50 (54.5\%)~\cite{kalluri2022memsac}, indicating that \textit{using an improved backbone may have the same effect as using a complex adaptation algorithm} on the target accuracy. While the general competence of ViT-backbones is well known, our study confirms that these improvements also extend to the case of out-of-domain robustness. 
We also observe that the relative ranking of different architectures widely varies across datasets, highlighting that the type of domain transfer influences domain robustness. 

\begin{figure*}[t]
\begin{center}

    \begin{minipage}[b]{0.75\textwidth}
        \centering
        \includegraphics[width=\textwidth]{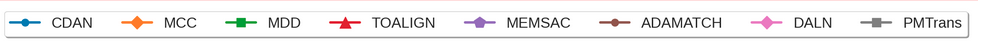}
    \end{minipage}
    
    \begin{minipage}[b]{0.42\textwidth}
        \centering
        \includegraphics[width=\textwidth]{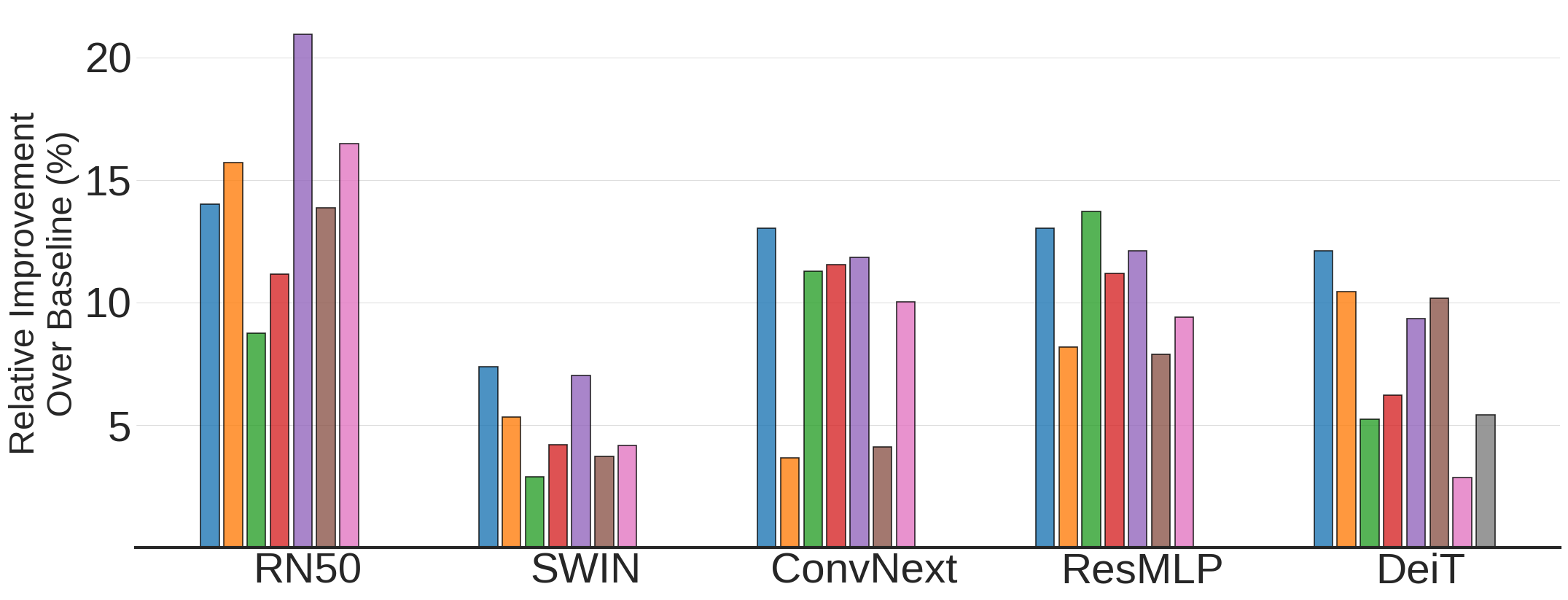}
        \vspace{-16pt}
        \captionsetup{width=\textwidth, font=footnotesize}
        \subcaption{{DomainNet (Real$\rightarrow$Clipart)} }
        \label{fig:arch_dnet_rc}
    \end{minipage}
    ~~~~~
     \begin{minipage}[b]{0.42\textwidth}
        \centering
        \includegraphics[width=\textwidth]{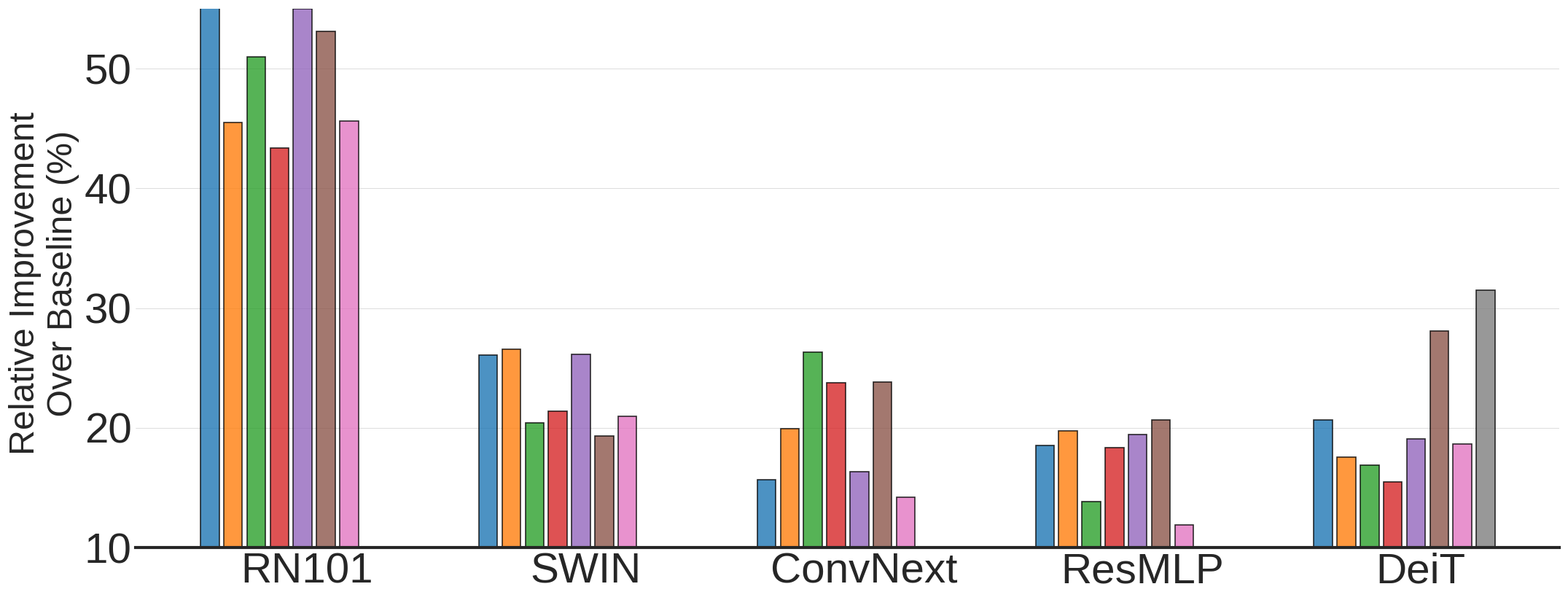}
        \vspace{-16pt}
        \captionsetup{width=\textwidth, font=footnotesize}
        \subcaption{{visDA (Synthetic$\rightarrow$Real)} }
        \label{fig:arch_dnet_visda}
    \end{minipage}
    
    \begin{minipage}[b]{0.42\textwidth}
        \centering
        \includegraphics[width=\textwidth]{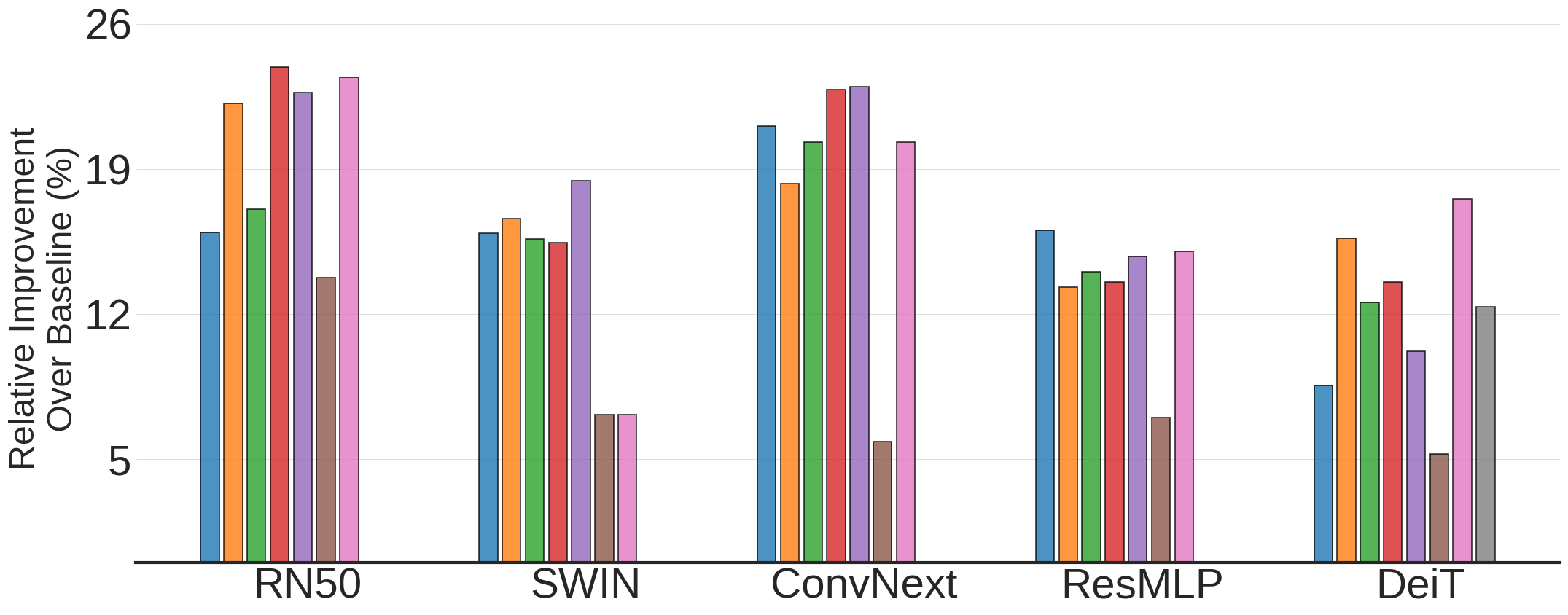}
        \vspace{-16pt}
        \captionsetup{width=\textwidth, font=footnotesize}
        \subcaption{{CUB200 (CUB$\rightarrow$Drawing)} }
        \label{fig:arch_cub_cd}
    \end{minipage}
    ~~~~~
    \begin{minipage}[b]{0.42\textwidth}
        \centering
        \includegraphics[width=\textwidth]{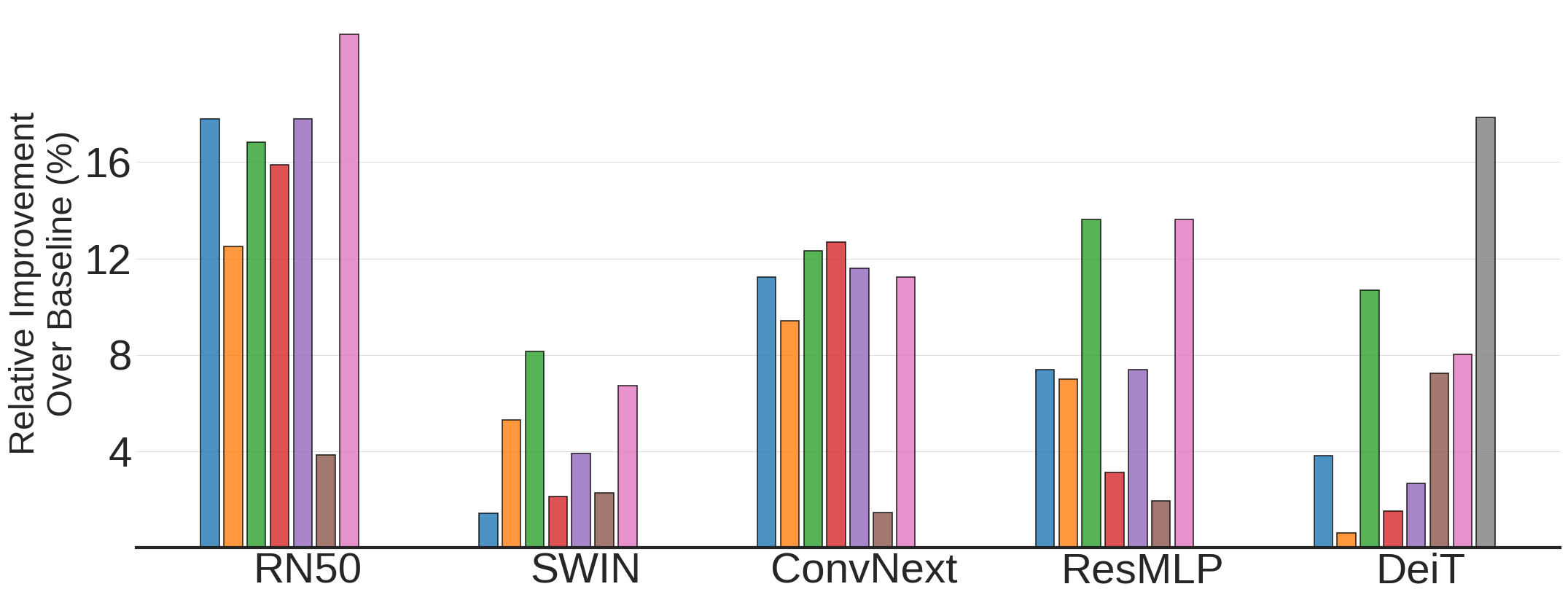}
        \vspace{-16pt}
        \captionsetup{width=\textwidth, font=footnotesize}
        \subcaption{{OfficeHome (Art$\rightarrow$Product)} }
        \label{fig:arch_ofh_ap}
    \end{minipage}
    \captionsetup{width=\textwidth, font=footnotesize}
    \caption{{\bf Better backbones diminish gains from UDA. } For each UDA method, we show the gain in accuracy relative to a baseline trained only using source-data. Across datasets, we observe that benefits offered by UDA approaches over the baseline diminish with backbones that have improved domain-robustness properties.} 
    \label{fig:arch_da_ablation}
    \vspace{-3em}
\end{center}
\end{figure*}

\vspace{2pt}
\newpara{UDA Gains Diminish With Newer Architectures}
We next ask the question if these benefits are complementary to the UDA method itself, and explore the viability of incorporating these advanced architectures into existing UDA methods. 
From \cref{fig:arch_da_ablation}, we observe that most methods do yield complimentary benefits over a source-only trained baseline even with newer architectures, but the \textit{relative improvement offered by UDA methods over this baseline tends to diminish when using better backbones}. Looking at the relative gain in accuracy over a source-only baseline, on Real$\rightarrow$Clipart in \cref{fig:arch_dnet_rc}, the best adaptation method provides 20\% relative gain over the baseline using ResNet-50, which falls to just 7\% with Swin and 10\% with DeiT backbone. Similarly, the relative gains offered by best UDA methods fall from 18\% with ResNet-50 to 8\% using Swin on Art$\rightarrow$Product in \cref{fig:arch_ofh_ap}. These observation also holds for visDA \cref{fig:arch_dnet_visda} and CUB200 \cref{fig:arch_cub_cd} datasets. The trends using the absolute accuracy drop also remain the same, while the relative drop further accounts for the strong source domain accuracy using advanced backbones. 
These results seem to suggest that the impact of many UDA methods is not really independent of the backbone used, and often tends to diminish in presence of better backbones which have better domain robustness properties. Furthermore, the \textit{relative ranking of the best adaptation method and backbone changes across datasets}, and is not consistent. For example, an older and simpler method like CDAN gives best accuracies in \cref{fig:arch_dnet_rc} with Swin, ConvNext and DeiT, while MCC outperforms other methods with a ResMLP backbone. 
We also show the more results on DomainNet and OfficeHome in \supp{}, and the results follow similar trends, where one of the more recent architectures significantly diminishes the returns yielded by all UDA methods. 

\newpara{Difference From Prior Works}
{While prior works like \cite{kim2022broad} only show this trend for classical UDA methods~\cite{CDAN, saito2018maximum} without using a standardized framework, we additionally show that this issue extends to more recent state-of-the-art UDA algorithms~\cite{wei2021toalign, Chen_2022_CVPR, kalluri2022memsac} as well, including methods using vision transformer backbones~\cite{zhu2023patchmix} using the proposed UDA-Bench, yielding several novel observations. For instance, we show in \cref{fig:arch_da_ablation} that the current SOTA method PMTrans~\cite{zhu2023patchmix} performs worse than DALN on CUB200 and CDAN on DomainNet when all of them use the same DeiT backbone, highlighting the key need to standardize backbones and architectures before comparing different methods. }




\subsection{How much unlabeled data can UDA methods use?}
\label{sec:unlabeled_data}

\newpara{Motivation} 
Although UDA holds great potential in leveraging unlabeled data from a target domain to enhance performance, an insight into their scalability properties in relation to the quantity of unlabeled data is lacking. These scaling properties are important to 
inform us which method has the greatest potential to improve performance when more unlabeled data becomes accessible, motivating us to study how much unlabeled data do UDA methods actually consume.

\newpara{Experimental Setup} To study the effects of data volume, we sample \{1, 5, 10, 25, 50, 100\}\% of the data from the target domain and run the adaptation algorithm using each of these subsets as the unlabeled data. We repeat the experiment with three different seeds in each case and report the mean accuracy to eliminate sampling bias. To avoid tail effects, we perform stratified sampling so that the label distribution is constant across all the subsets. 
Specifically, we sample $x\%$ of data from each category individually which helps to preserve the tail properties of the resulting sub-sampled dataset. We also make sure that all categories have at least 1 image in the sub-sampled dataset. Note that the label information in the target is used only during sampling, but not during training. 
We note the possibility of hyper-parameter sensitivity to the amount of target unlabeled data, but do not preform any additional tuning to keep the number of experiments manageable. We restrict to using DomainNet and VisDA in our analysis as those are the largest available datasets for domain adaptation, and show results using another recent large-scale adaptation benchmark GeoNet in the \supp{}.
The already tiny data volume in OfficeHome and CUB200 prevents their use in a scalability study like this. 

\begin{figure*}[t]
\begin{center}
    \begin{minipage}[b]{0.75\textwidth}
        \centering
        \includegraphics[width=\textwidth]{figures/arch_plot_legend_with_daln.png}
    \end{minipage}
    
    \begin{minipage}[b]{0.4\textwidth}
        \centering
        \includegraphics[width=\textwidth]{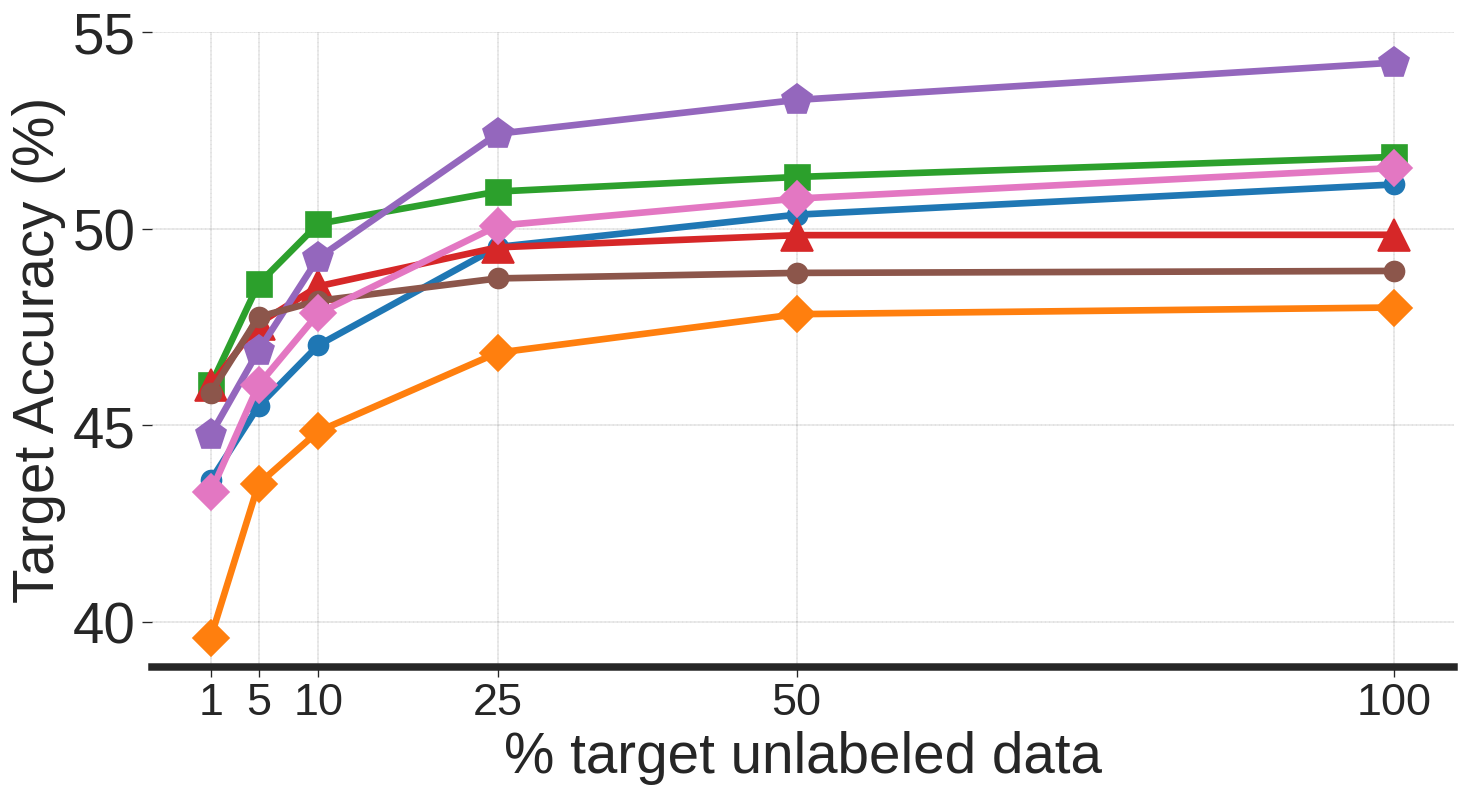}
        \captionsetup{width=\textwidth, font=footnotesize}
        \subcaption{Real$\rightarrow$Clipart (Resnet-50)}
        \label{fig:datavol_rc}
    \end{minipage}
    ~~~~~~
    \begin{minipage}[b]{0.4\textwidth}
        \centering
        \includegraphics[width=\textwidth]{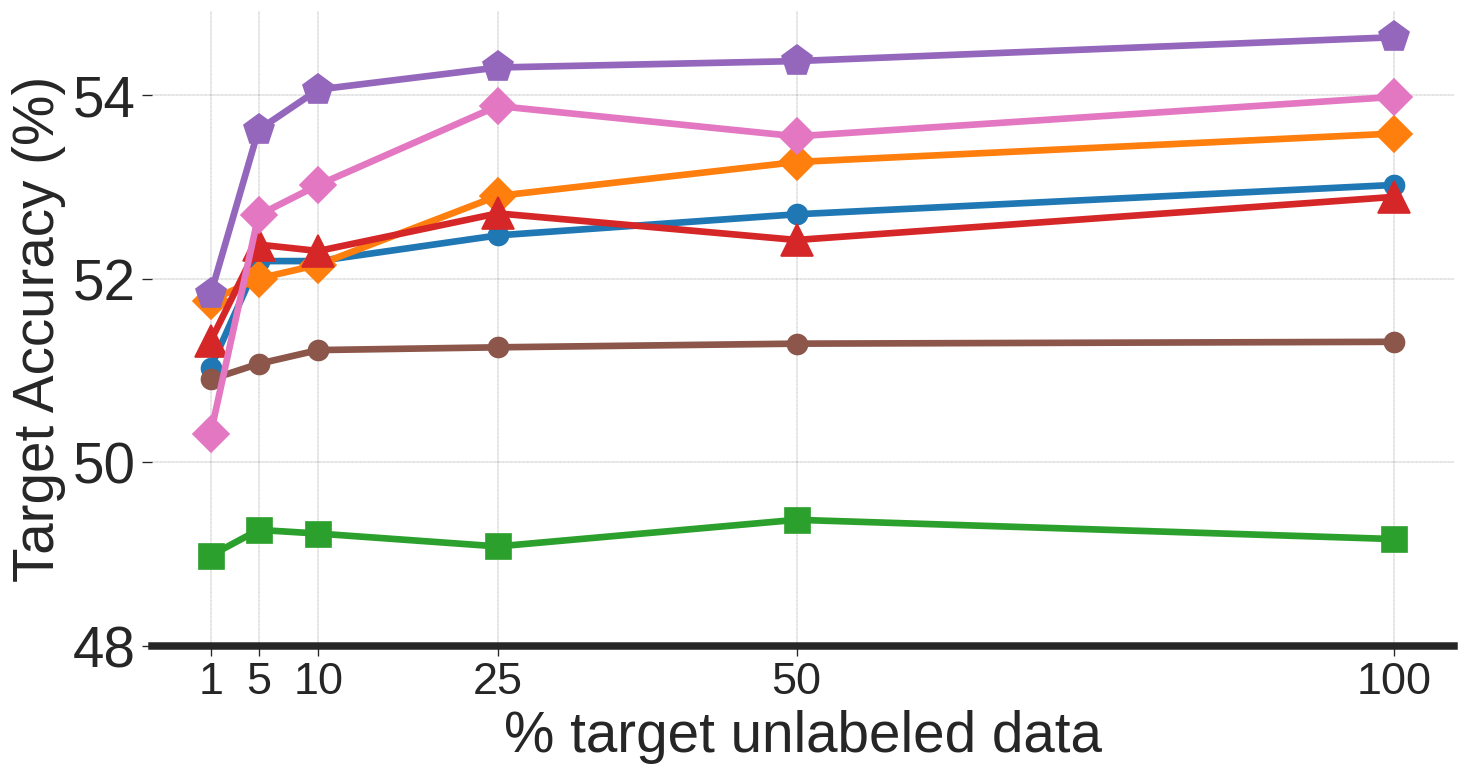}
        \captionsetup{width=\textwidth, font=footnotesize}
        \subcaption{Painting$\rightarrow$Real (Resnet-50)}
        \label{fig:datavol_pr}
    \end{minipage}

    \begin{minipage}[b]{0.4\textwidth}
        \centering
        \includegraphics[width=\textwidth]{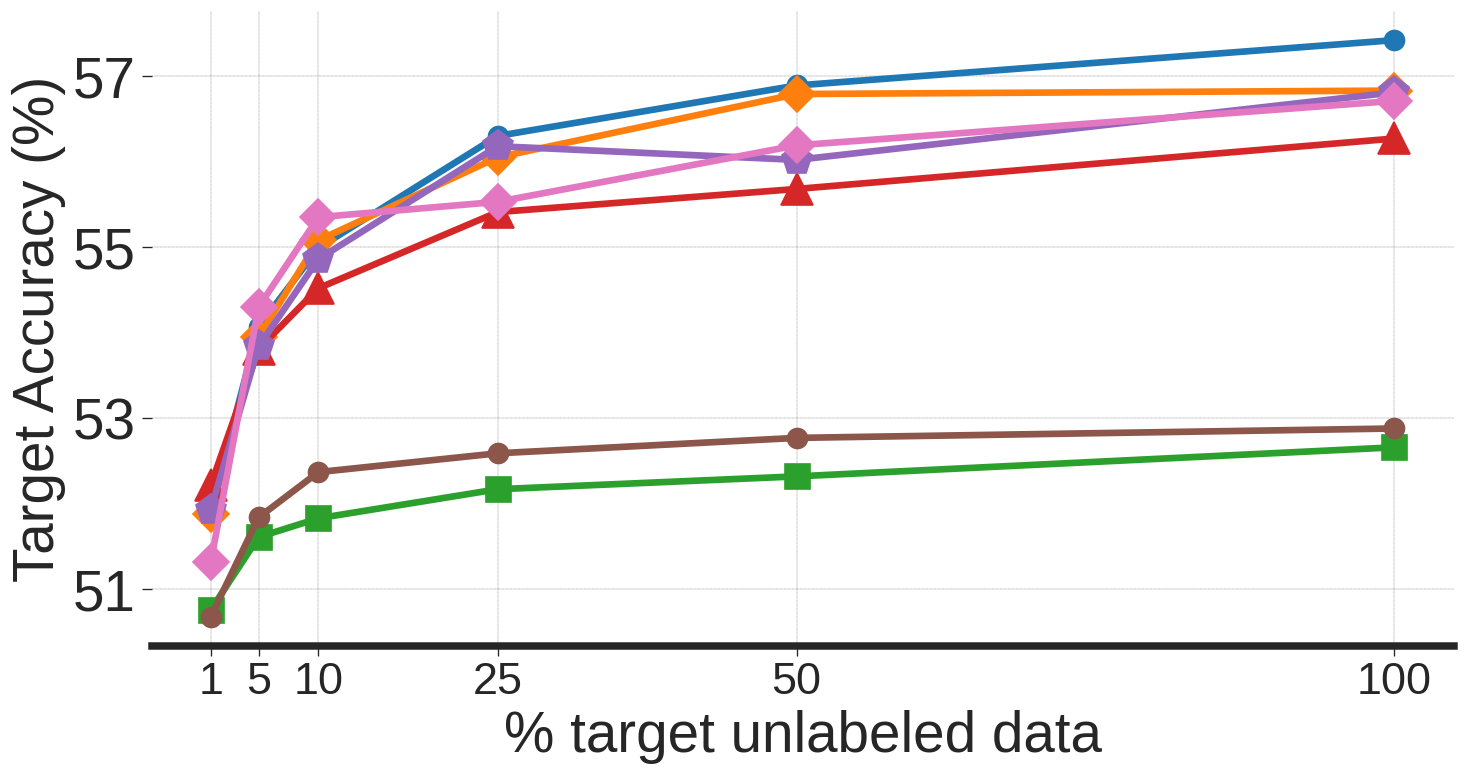}
        \captionsetup{width=\textwidth, font=footnotesize}
        \subcaption{Real$\rightarrow$Clipart (ConvNext-t)}
        \label{fig:datavol_rc_convnext}
    \end{minipage}
    ~~~~~~
    \begin{minipage}[b]{0.4\textwidth}
        \centering
        \includegraphics[width=\textwidth]{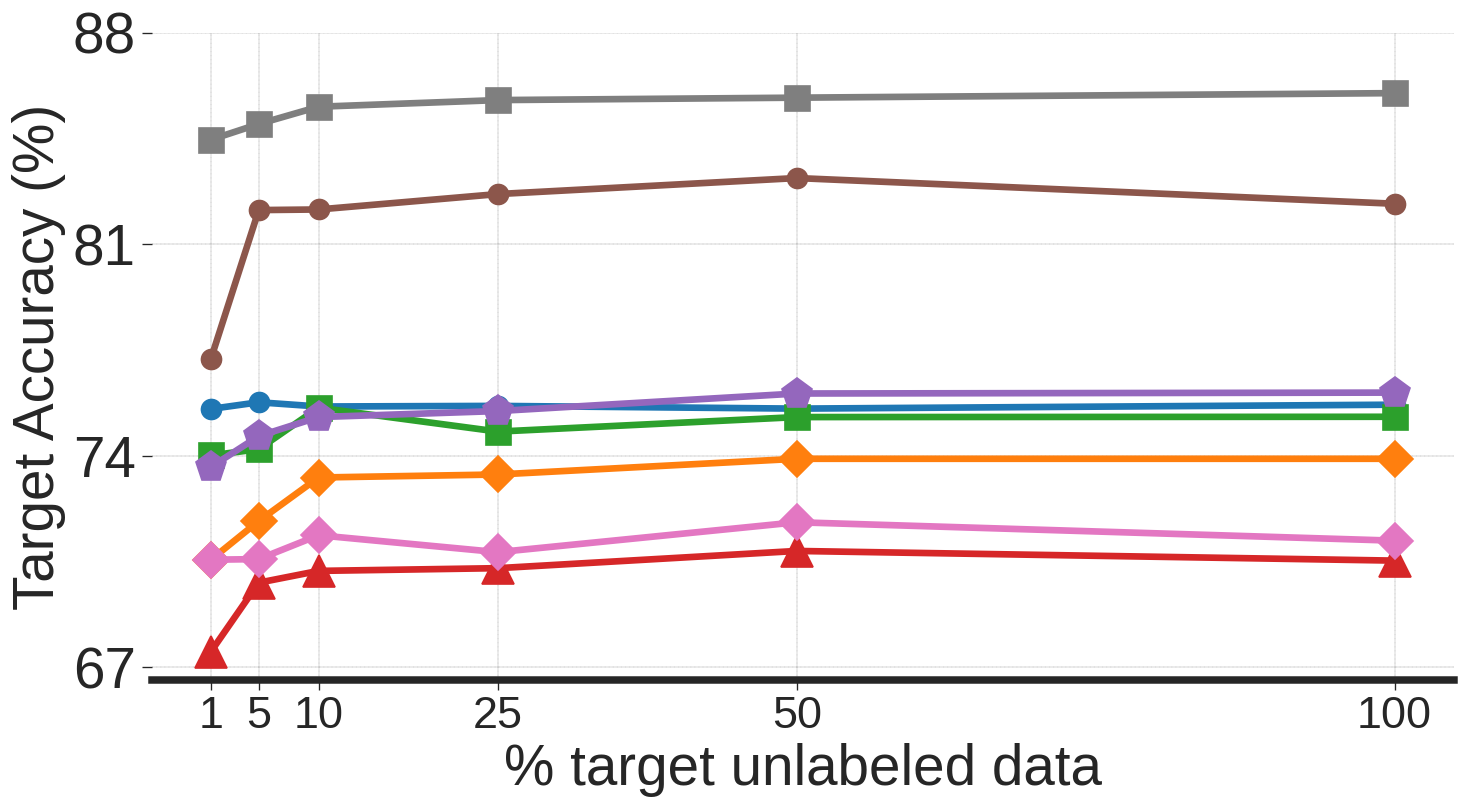}
        \captionsetup{width=\textwidth, font=footnotesize}
        \subcaption{VisDA (DeiT)}
        \label{fig:datavol_visda}
    \end{minipage}
    
    \captionsetup{width=\textwidth, font=footnotesize}
    \caption{{\bf How much unlabeled data can UDA methods use? } Across different adaptation datasets and backbones (Resnet50 in \subref{fig:datavol_rc}, \subref{fig:datavol_pr}, ConvNext in \subref{fig:datavol_rc_convnext} and DeiT in \subref{fig:datavol_visda}), we find that the performance of several UDA methods saturates quickly with respect to amount of target data, showing their limited efficiency in utilizing the unlabeled samples. In most cases, using only 25\% of the data results in $<1\%$ drop in accuracy.} 
    \label{fig:datavol_ablation}
    \vspace{-3em}
\end{center}
\end{figure*}

\newpara{UDA Accuracy Does Not Increase With More Unlabeled Data.}
Remarkably the trends from \cref{fig:datavol_ablation} indicate that on all the settings \textit{the accuracy achieved by the unsupervised adaptation saturates rather quickly with respect to the unlabeled data}. This trend holds for almost all of the studied adaptation methods, including adversarial~\cite{CDAN}, non-adversarial~\cite{berthelot2021adamatch}, consistency based~\cite{kalluri2022memsac} and pseudo-label based~\cite{zhu2023patchmix} methods. The gains remain less than $2\%$ in most cases even when scaling unlabeled data four-fold (from 25\% to 100\%). For example, on R$\rightarrow$C (\cref{fig:datavol_rc}), the accuracy achieved at using just $25\%$ of the unlabeled data is within $1\%$ of the accuracy obtained at $100\%$ of the data using any adaptation method. In P$\rightarrow$R, (\cref{fig:datavol_pr}) the accuracy plateaus much earlier, at around $10-15\%$ of the unlabeled data. 
Similar results are observed using a different backbone like DeiT with a purely transformer-based method PMTrans~\cite{zhu2023patchmix} (\cref{fig:datavol_visda}), where the performance saturates after using only $10\%$ of the unlabeled data. These results suggest that even in cases where abundant unlabeled data becomes available, current UDA methods cannot leverage the potential benefits of this data to enhance performance. We also show more results on DomainNet in \supp, where the observations follow similar trends. 

Furthermore, we juxtapose this observation with a similar ablation using source labeled data in the \supp{}, and identify that source supervision has a more pronounced effect on the target accuracy than target unlabeled data. Specifically, increasing source labeled data from 50\% to 100\% results in upto 10\% gain in target accuracy (as opposed to $<1\%$ observed using similar scale increase in target unlabeled data). 


%

\begin{figure*}[!t]
\begin{center}
    \begin{minipage}[b]{\textwidth}
        \centering
        \hfill \includegraphics[width=0.64\textwidth]{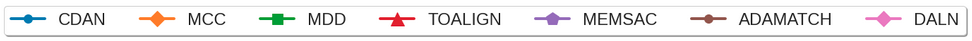}
    \end{minipage}

    \begin{minipage}[b]{0.29\textwidth}
        \centering
        \includegraphics[width=\textwidth]{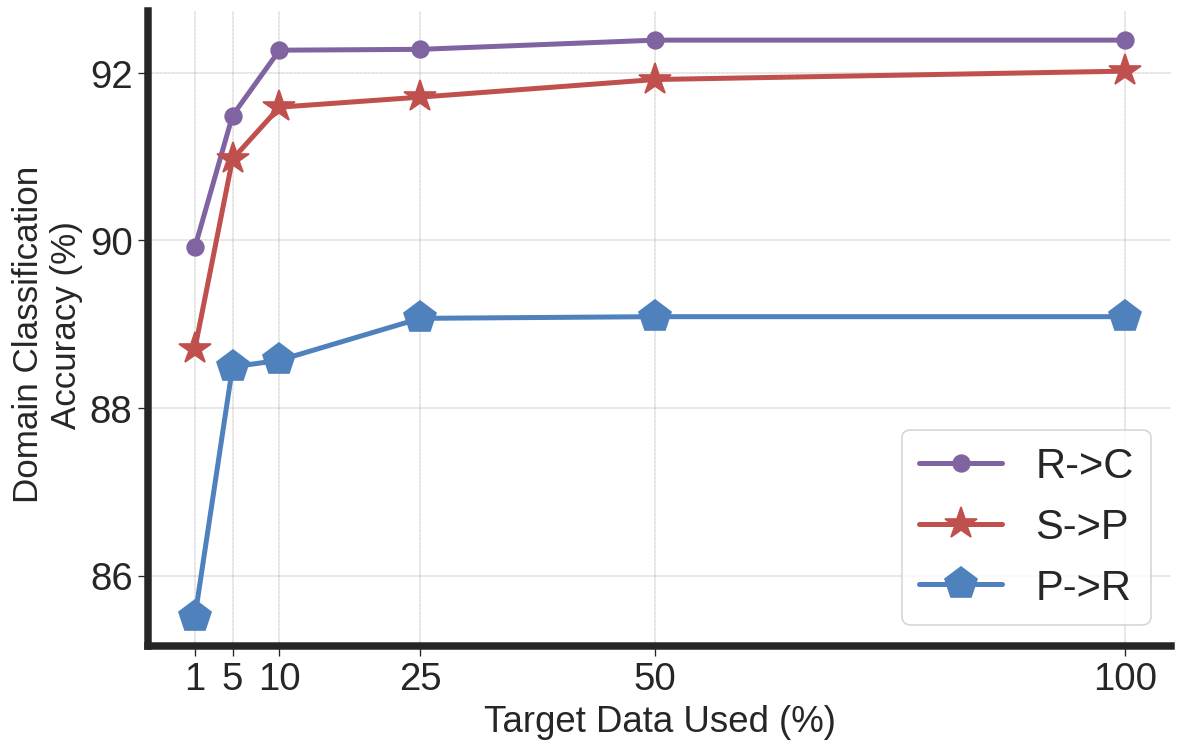}
        \captionsetup{width=\textwidth, font=footnotesize}
        \subcaption{Accuracy Saturation}
        \label{fig:domain_classification}
    \end{minipage}
    \hfill
    \begin{minipage}[b]{0.32\textwidth}
        \centering
        \includegraphics[width=\textwidth]{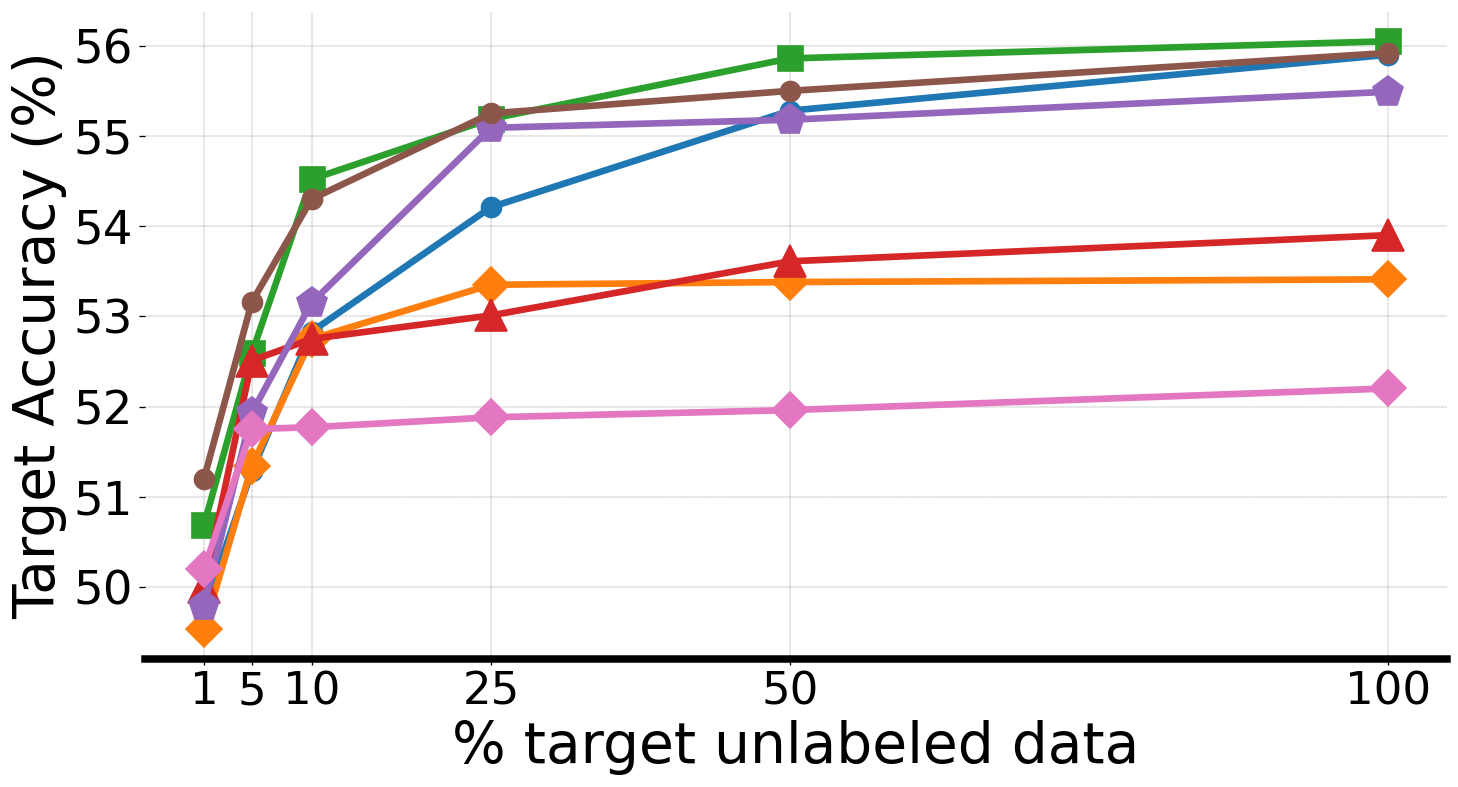}
        \captionsetup{width=\textwidth, font=footnotesize}
        \subcaption{Random Sampling}
        \label{fig:random_sample}
    \end{minipage}
    \hfill
     \begin{minipage}[b]{0.32\textwidth}
        \centering
        \includegraphics[width=\textwidth]{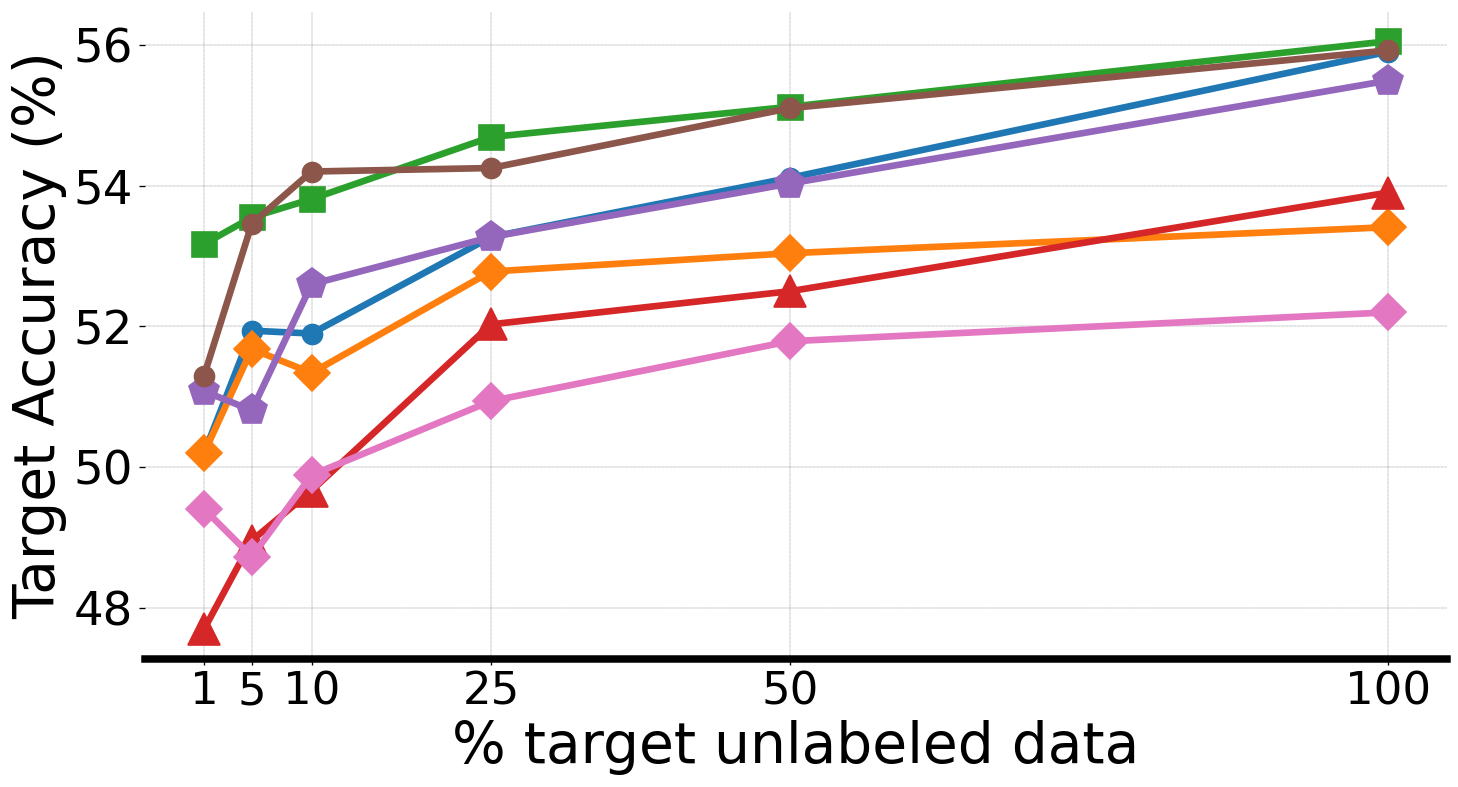}
        \captionsetup{width=\textwidth, font=footnotesize}
        \subcaption{Class-split Sampling}
        \label{fig:split_sample}
    \end{minipage}
    
    \captionsetup{width=\textwidth, font=footnotesize}
    \caption{(\subref{fig:domain_classification}){\bf Saturation of the domain classification accuracy} is observed even with small amount of unlabeled data, potentially explaining the poor sample efficiency of UDA methods employing adversarial domain alignment. (\subref{fig:random_sample},\subref{fig:split_sample}) {\bf Role of the sampling technique adopted} We study the behavior of UDA methods with respect to target unlabeled data using two additional sampling techniques: random sampling in (\subref{fig:random_sample}) and split-class sampling in (\subref{fig:split_sample}). Our observation that UDA methods under-utilize unlabeled data holds for both of these cases as well.} 
    \label{fig:sampling_technique}
    \vspace{-16pt}
\end{center}
\end{figure*}

\newpara{Investigating Poor Data Efficiency of UDA Methods}
We hypothesize that the main reason behind poor unlabeled sample efficiency is the underlying adaptation objective employed, which fails to effectively utilize growing amounts of unlabeled data. As an example, we take the objective of domain classification, which forms the backbone of several adversarial UDA methods~\cite{DANN, CDAN}, and examine its data efficiency.  
We plot the accuracy of the domain discrimination objective itself against the quantity of unlabeled samples in \cref{fig:domain_classification} for different settings from DomainNet. We notice that the domain classification accuracy reaches a plateau after using approximately 25\% of the data, potentially explaining the saturation of the adaptation accuracy in methods that rely on this objective for bridging the domain gap. While this explains adversarial alignment based methods, we posit that similar limitations impact other types of adaptation approaches including self-training, pseudo-label or consistency-based methods. 

\newpara{UDA Empirical Data Efficiency Does Not Match Theory.}
The above observation stands in stark contrast to the theoretical framework of domain adaptation established by Ben-David et al.\cite{ben2010theory}, which underpins several UDA methods. Their theoretical analysis suggests an inverse relationship between target sample size and target error (Theorem 2 from \cite{ben2010theory}), further highlighting the importance of empirical study like ours using a unified framework like UDA-Bench to understand the bridge between theory and practice. Our observation from UDA is also different from prior scalability studies in supervised~\cite{sun2017revisiting}, weakly-supervised~\cite{singh2022revisiting} and self-supervised learning~\cite{goyal2019scaling} literature, where increasing labeled or unlabeled data significantly enhances performance.


\newpara{Similar Results Hold For Other Sampling Techniques} {In addition to the class-balanced sampling procedure in \cref{fig:datavol_ablation}, we also show results using two other sampling techniques, random sampling and split-class sampling in \cref{fig:random_sample} and \cref{fig:split_sample} respectively. In \cref{fig:random_sample}, we randomly select $x\%$ of images from the whole dataset without any class-aware sampling, and show the general observation that UDA methods reach a performance plateau after utilizing a limited amount of unlabeled data holds, where using only 50\% of the unlabeled data resulted no drop in performance for most of the methods.
In \cref{fig:split_sample}, we adopt a \textit{split-class} sampling technique, where we first randomly select half the classes, and remove $2x\%$ of data from these classes while keeping images from the rest of the classes the same. This sampling technique would reveal insights into scenarios where the tail properties of the category distribution exhibit significant skewness, and adding unlabeled data translates to correcting the skewed tail property of the dataset. However, the gains yielded from adding more unlabeled data is still limited. Even when the overall trends look positive with non-saturated performance, the absolute gain \textit{is still less than 2\%} while doubling the amount of unlabeled data from 50\% to 100\%, matching the observations made with other sampling techniques. }

\subsection{Does pre-training data matter in UDA?} 
\label{sec:pretraining}

\newpara{Motivation} 
Following recent works that reveal the importance of pre-training data in influencing downstream accuracy~\cite{cole2022does}, we revisit a standard practice in UDA to adopt ImageNet pre-trained backbone irrespective of the downstream adaptation task.
While Kim et al.~\cite{kim2022broad} share similar motivations as ours, a notable distinction lies in their focus on \textit{scaling} pre-training data and architectures, while we offer complementary insights by exploring the relationship between the \textit{type} of pre-training and downstream adaptation maintaining a \textit{constant} datasize.

\begin{table*}[t]
    \centering
    \captionsetup{width=\textwidth, font=footnotesize}
    \caption{{\bf In-task Supervised pre-training helps domain adaptation.} We analyze the relationship between data used for supervised pre-training and downstream adaptation for source-only transfer as well as several UDA methods including MemSAC~\cite{kalluri2022memsac}, ToAlign~\cite{wei2021toalign}, MDD~\cite{zhang2019bridging} and DALN~\cite{Chen_2022_CVPR}. We show that \textit{in-task} supervised pre-training significantly helps adaptation. All models use ResNet-50 backbone. IN:ImageNet, PL:Places-205, NAT:iNaturalist.}
    \label{tab:pt_sup}
  \begin{minipage}[!thp]{0.9\textwidth}
    \resizebox{\textwidth}{!}{%
    \begin{tabular}{c cccc c ccc c ccc c ccc c ccc}
    \toprule
       && \multicolumn{3}{c}{Plain Transfer (\footnotesize{no adapt})} && \multicolumn{3}{c}{ToAlign~\cite{wei2021toalign}} && \multicolumn{3}{c}{MemSAC~\cite{kalluri2022memsac}} && \multicolumn{3}{c}{MDD~\cite{zhang2019bridging}} && \multicolumn{3}{c}{DALN~\cite{Chen_2022_CVPR}} \\
      \cmidrule{3-5} \cmidrule{7-9} \cmidrule{11-13} \cmidrule{15-17} \cmidrule{19-21}
      Pre-training && DNet & GeoP & CUB && DNet & GeoP & CUB && DNet & GeoP & CUB && DNet & GeoP & CUB && DNet & GeoP & CUB  \\
      \cmidrule{1-1} \cmidrule{3-5} \cmidrule{7-9} \cmidrule{11-13} \cmidrule{15-17} \cmidrule{19-21}
      IN-1M && \textbf{41.46} & 34.55 & 50.20  && \textbf{49.29} & 30.42 & 62.78 && \textbf{50.75} & 32.98 & 62.92 && 
      \textbf{42.40} & 30.84 & 59.84 && \textbf{47.59} & 26.85 & 61.45 \\
      PL-1M && 35.14 & \textbf{41.95} & 40.83 && 38.55 & \textbf{34.9} & 55.29 && 41.93 & \textbf{40.16} & 54.22 && 
      34.94 & \textbf{37.90} & 51.14 && 39.21 & \textbf{36.23} & 50.74 \\
      NAT-1M && 33.77 & 31.53 & \textbf{58.77} && 37.65 & 26.81 & \textbf{67.47} && 38.67 & 29.99 & \textbf{67.34} && 
      32.29 & 26.79 & \textbf{63.72} && 37.30 & 24.69 & \textbf{66.80} \\
      \bottomrule
    \end{tabular}
    }
  \end{minipage}
    \vspace{-8pt}
\end{table*}

\newpara{Experimental Setup} We use ImageNet~\cite{ILSVRC15}, Places-205~\cite{zhou2017places} and  iNaturalist-2021~\cite{van2021benchmarking} as datasets during pre-training.
While ImageNet contains images from diverse natural and object categories, Places-205 is designed for scene classification and iNaturalist contains images of bird species.
We select 1M images each from ImageNet, Places-205 and iNaturalist datasets (indicated as IN-1M, PL-1M and NAT-1M respectively) to keep the size of the pre-training datasets constant, allowing us to decouple the impact of nature of data from the volume of the dataset. In terms of pre-training methods, we use supervised pre-training using labeled data, along with recent state-of-the-art self-supervised methods SwAV~\cite{caron2020unsupervised}, MoCo-V3~\cite{chen2021empirical} and MAE~\cite{he2022masked}, which broadly cover the three families of clustering, contrastive and masked auto-encoding based methods for self-supervised learning. We train SwAV on ResNet-50, MoCo on ViT-S/16 and MAE on ViT-B/16 architectures, along with supervised pre-training on ResNet-50, thereby extending our inferences to a diverse pool of pretraining data and architectures.
%
For the downstream adaptation tasks, we use Real$\rightarrow$Clipart on DomainNet, CUB$\rightarrow$Drawing on CUB200 and USA$\rightarrow$Asia on GeoPlaces covering three distinct application scenarios for adaptation on objects, birds and scenes respectively. 
To prevent overlap between pre-training and adaptation data, we remove images from Places-205 that are also present in GeoPlaces and remove images from iNaturalist that belong to the same class as those in CUB200. 

%
%
%
\newpara{Supervised Pre-training Using In-Task Data Helps UDA} 
In our analysis, we loosely consider pre-training on ImageNet, iNaturalist and Places205 to be \emph{in-task pre-training} for downstream adaptation on DomainNet, CUB200 and GeoPlaces respectively due to the matching style of images. We show our results using supervised pre-training on Resnet-50 in \cref{tab:pt_sup} for plain source-only transfer (no adaptation), as well as  adaptation using ToAlign, MemSAC, MDD and DALN. Across the board, we observe that \textit{in-task pre-training always yields better results on downstream adaptation} even when using the same amount of data. 
Focusing on plain transfer from \cref{tab:pt_sup}, the de-facto choice of ImageNet pre-training gives 50.2\% on CUB$\rightarrow$Drawing transfer task, while just switching the pre-training dataset to iNaturalist2021 yields 58.7\% accuracy with an absolute improvement of 8.5\%. 
Likewise, we observe a non-trivial improvement of 7.4\% absolute accuracy for GeoPlaces (34.5\% to 41.9\%) using Places205 for pre-training even without any adaptation, challenging the common assumption of using an ImageNet-pretrained model irrespective of the downstream task. 
%
We hypothesize that supervised pre-training on in-task data creates strong priors with more relevant features, thereby enhancing generalization on similar downstream tasks. Consequently, we conclude that selecting in-task pre-trained models is a viable approach to improve accuracy, particularly when target unlabeled data is unavailable.
{While similar observations have been made before in continual pre-training~\cite{reed2022self} or language models~\cite{DBLP:conf/acl/GururanganMSLBD20}, our difference lies in highlighting this behavior for the specific case of UDA through a unified framework and controlled empirical study. }

\begin{table*}[!t]
    \captionsetup{width=\textwidth, font=footnotesize}
    \caption{{\bf Self-supervised pre-training and domain adaptation.} We find that self-supervised pre-training on object-centric images (on ImageNet) help downstream accuracy on object-centric adaptation (on DomainNet and CUB200), while scene-centric pre-training (on Places205) benefit adaptation on scene-centric GeoPlaces task. IN:ImageNet, PL:Places-205, NAT:iNaturalist}
    \label{tab:pt_ssl}
  \begin{minipage}[t]{0.48\textwidth}
    \centering
    \resizebox{\textwidth}{!}{%
    \begin{tabular}{lcccc c ccc c ccc}
    \toprule
       && \multicolumn{3}{c}{SwAV (ResNet50)~\cite{caron2020unsupervised}} && \multicolumn{3}{c}{MoCo-V3 (ViT-s/16)~\cite{chen2021empirical}} && \multicolumn{3}{c}{MAE (ViT-b/16)~\cite{he2022masked}} \\
      \cmidrule{3-5} \cmidrule{7-9} \cmidrule{11-13}
      Pretraining && DNet & GeoP & CUB && DNet & GeoP & CUB && DNet & GeoP & CUB  \\
      \cmidrule{1-1} \cmidrule{3-5} \cmidrule{7-9} \cmidrule{11-13}
      IN-1M && \textbf{36.51} & 35.76 & \textbf{31.59} && \textbf{30.48} & 31.13 & \textbf{40.7} && \textbf{38.58} & 35.85 & \textbf{52.34} \\
      PL-1M && 30.86 & \textbf{42.26} & 27.44 && 27.45 & \textbf{35.89} & 39.49 && 34.76 & \textbf{38.1} & 45.25 \\
      NAT-1M && 28.01 & 29.01 & 30.12 && 25.66 & 27.82 & 40.03 && 33.78 & 31.68 & 49.4 \\
      \bottomrule
    \end{tabular}
    }
    \subcaption{{\bf Plain Transfer (No Adaptation)}}
    \label{tab:pt_ssl_plain}
  \end{minipage}
    \hfill
  \begin{minipage}[t]{0.48\textwidth}
    \centering
    \resizebox{\textwidth}{!}{%
    \begin{tabular}{lcccccccccccc}
    \toprule
       && \multicolumn{3}{c}{SwAV (ResNet50)~\cite{caron2020unsupervised}} && \multicolumn{3}{c}{MoCo-V3 (ViT-s/16)~\cite{chen2021empirical}} && \multicolumn{3}{c}{MAE (ViT-b/16)~\cite{he2022masked}} \\
      \cmidrule{3-5} \cmidrule{7-9} \cmidrule{11-13}
      Pretraining && DNet & GeoP & CUB && DNet & GeoP & CUB && DNet & GeoP & CUB  \\
      \cmidrule{1-1} \cmidrule{3-5} \cmidrule{7-9} \cmidrule{11-13}
      IN-1M && \textbf{44.6} & 36.33 & \textbf{51.81} && \textbf{34.33} & 30.35 & \textbf{52.61} && \textbf{44.91} & 34.07 & \textbf{64.26} \\
      PL-1M && 36.48 & \textbf{41.14} & 39.49 && 30.83 & \textbf{35.51} & 46.99 && 39.56 & \textbf{37.00} & 53.68 \\
      NAT-1M && 31.6 & 28.75 & 45.65 && 28.24 & 26.01 & 48.46 && 38.48 & 28.74 & 59.7 \\
      \bottomrule
    \end{tabular}
    }
    \subcaption{{\bf Using MemSAC Adaptation}}
    \label{tab:pt_ssl_memsac}
  \end{minipage}
  \vspace{-16pt}
  
\end{table*}

\newpara{In-Task Pre-training is complementary to UDA method} 
We also observe that these benefits obtained from in-task supervised pre-training complement the advantages potentially obtained using UDA methods, resulting in additional improvements in accuracy.
From \cref{tab:pt_sup}, on CUB200, we observe $17.1\%$ and $17.3\%$ improvement using MemSAC and ToAlign respectively together with in-task pre-training, over standard practice of ImageNet-pretraining and fine-tuning on source data ($12\%$ from changing the backbone and further $5\%$ from the adaptation), \textit{setting a new state-of-the-art on CUB200 dataset} using in-task pre-training. On the other hand, a significant mismatch between the pre-training dataset and the downstream domain adaptation dataset (such as Places and Birds datasets), noticeably reduces the accuracy by ${>}10\%$ in most cases, underlining the dependence of model's generalization ability to the pre-training data.
%
While these findings may seem intuitive, it is important to note that all UDA methods consistently utilize ImageNet pre-training as the default, irrespective of the adaptation dataset. This may lead to practitioners assuming ImageNet pre-training as the optimal choice, potentially overlooking performance gains achievable by employing alternative pre-trained models tailored to the target task, as demonstrated by our empirical study.

\newpara{Nature of Pre-training Images matter for Self-supervised Learning} We show results for self-supervised setting in \cref{tab:pt_ssl}. 
We first note that supervised pre-training (\cref{tab:pt_sup}) achieves much higher accuracies after downstream adaptation compared to self-supervised pre-training. 
This is expected, as supervised pre-training captures richer object semantics through labels inherently benefiting any downstream task, while self-supervised learning relies on pretext tasks that may not impart equivalent semantic understanding.
In terms of pre-training data, we observe that both CUB200 and DomainNet benefit from self-supervised pre-training on ImageNet, while GeoPlaces still benefits from pre-training on Places205. This observation holds for both source-only transfer (\cref{tab:pt_ssl_plain}) as well as adaptation using MemSAC (\cref{tab:pt_ssl_memsac}). 
We posit that in a self-supervised setting, \textit{the nature of images in the datasets (whether object-centric or scene-centric) plays a crucial role in downstream transfer}. Specifically, unsupervised pre-training on object-centric images from ImageNet leads to improved image classification accuracies on DomainNet and CUB200. Conversely, unsupervised pre-training on scene-centric Places205 showcase better transfer performance in place recognition tasks on the GeoPlaces dataset. Among the two object-centric datasets, we find that the diversity of images in ImageNet is better for effective transfer compared to specific domain-based datasets like iNaturalist, as also highlighted in prior works for self-supervised learning~\cite{cole2022does}. Furthermore, this property is consistent across different kinds of self-supervised pretext tasks like SwAV, MoCo and MAE. 

\section{Conclusion}
In this work, we provide a holistic analysis of factors that impact the effectiveness UDA methods developed for image-classification, most of which are not apparent from standard training and evaluation practices. Through our innovation called UDA-bench that facilitates fair comparisons across UDA methods, we perform a controlled empirical study revealing key insights regarding the sensitivity of these methods to the backbone architecture, their limited efficiency in utilizing unlabeled data, and the potential for enhancing performance through in-task pre-training - where existing UDA theory proves highly inadequate for explaining several of our novel empirical observations.
In terms of limitations of the study, we only consider UDA designed for classification in this work, and our findings might or might not hold for other problem areas such as domain adaptive semantic segmentation. We also acknowledge the potential existence of other unexplored factors that may impact the performance of UDA methods beyond those studied here, and offer UDA-Bench as a suitable avenue for future research in this direction. 
Further, we mainly focus on the standard setting in unsupervised adaptation, but believe that a deeper understanding of algorithms in such conventional settings forms the backbone for future studies in other variants including source-free~\cite{liang2020we}, semi-supervised~\cite{saito2019semi} and universal~\cite{you2019universal} DA methods. Several other avenues like adaptation of vision-language models~\cite{radford2021learning, zhai2023sigmoid} and emerging generative models~\cite{rombach2022high, liu2023improvedllava} are also left to a future work. 

\section*{Acknowledgements}

We acknowledge support from NSF and a Google Award for Inclusion Research.

\bibliographystyle{splncs04}
\bibliography{main}

\newpage
\appendix
\section{UDABench: Code Overview}
\label{sec:code_overview}
 
We build our codebase using PyTorch following several open-source deep-learning libraries like Detectron~\cite{wu2019detectron2} and PyTorch3D~\cite{ravi2020pytorch3d}. The overarching motivation in designing UDA-Bench is to standardize evaluation and training of existing unsupervised adaptation methods to facilitate fair comparative studies like ours, while also enabling quick prototyping and design of new adaptation methods in the future. 
UDA-Bench is designed to be flexible to incorporate newer architecture backbones, classifier modules, optimizers, loss functions, dataloaders and training methods with minimal effort and design overhead, allowing researchers to build upon existing adaptation methods to develop new innovations in unsupervised adaptation. 

We re-implement several classical as well as state-of-the-art UDA methods in UDA-Bench. We keep the adaptation independent hyper-parameters (such as architectures, batch sizes) same across the methods, and use the adaptation-specific hyper-parameters as recommended in the respective methods. We use the open-source repositories of prior UDA methods from the links given below. 
\begin{itemize}
    \item {\bf CDAN:} \href{https://github.com/thuml/CDAN/tree/master}{https://github.com/thuml/CDAN/tree/master}
    \item {\bf MCC:} \href{https://github.com/thuml/Versatile-Domain-Adaptation}{https://github.com/thuml/Versatile-Domain-Adaptation}
    \item {\bf MDD:} \href{https://github.com/thuml/MDD}{https://github.com/thuml/MDD}
    \item {\bf ToAlign:} \href{https://github.com/microsoft/UDA}{https://github.com/microsoft/UDA}
    \item {\bf MemSAC:} \href{https://github.com/ViLab-UCSD/MemSAC_ECCV2022}{https://github.com/ViLab-UCSD/MemSAC\_ECCV2022}
    \item {\bf AdaMatch:} \href{https://github.com/google-research/adamatch}{https://github.com/google-research/adamatch}
    \item {\bf DALN:} \href{https://github.com/xiaoachen98/DALN}{https://github.com/xiaoachen98/DALN}
    \item {\bf PMTrans:} \href{https://github.com/JinjingZhu/PMTrans}{https://github.com/JinjingZhu/PMTrans}
\end{itemize}

\section{Additional Results on DomainNet and OfficeHome} 

\begin{figure*}[!t]
\begin{center}

    \begin{minipage}[b]{0.75\textwidth}
        \centering
        \includegraphics[width=\textwidth]{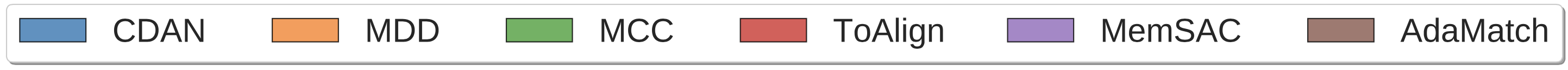}
    \end{minipage}

    \begin{minipage}[b]{0.46\textwidth}
        \centering
        \includegraphics[width=\textwidth]{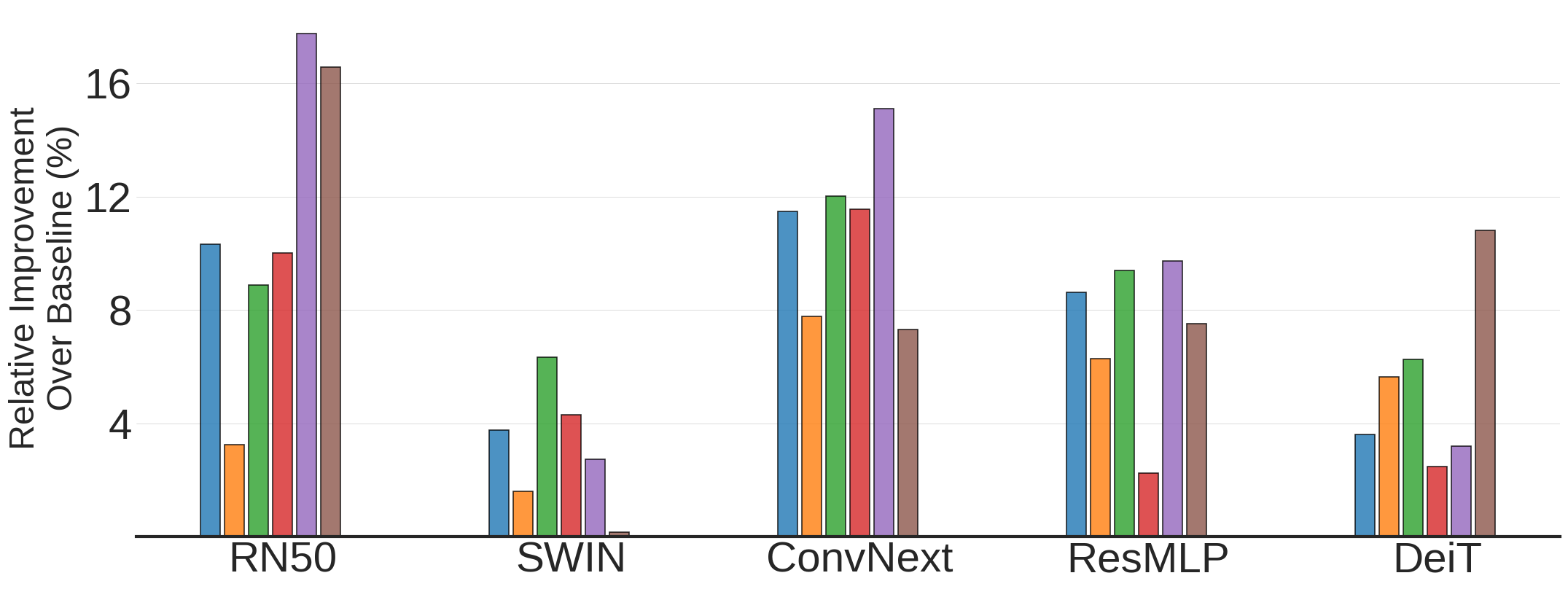}
        \vspace{-16pt}
        \captionsetup{width=\textwidth, font=footnotesize}
        \subcaption{{DomainNet} }
        \label{fig:arch_dnet_cs}
    \end{minipage}
    \hfill
    \begin{minipage}[b]{0.46\textwidth}
        \centering
        \includegraphics[width=\textwidth]{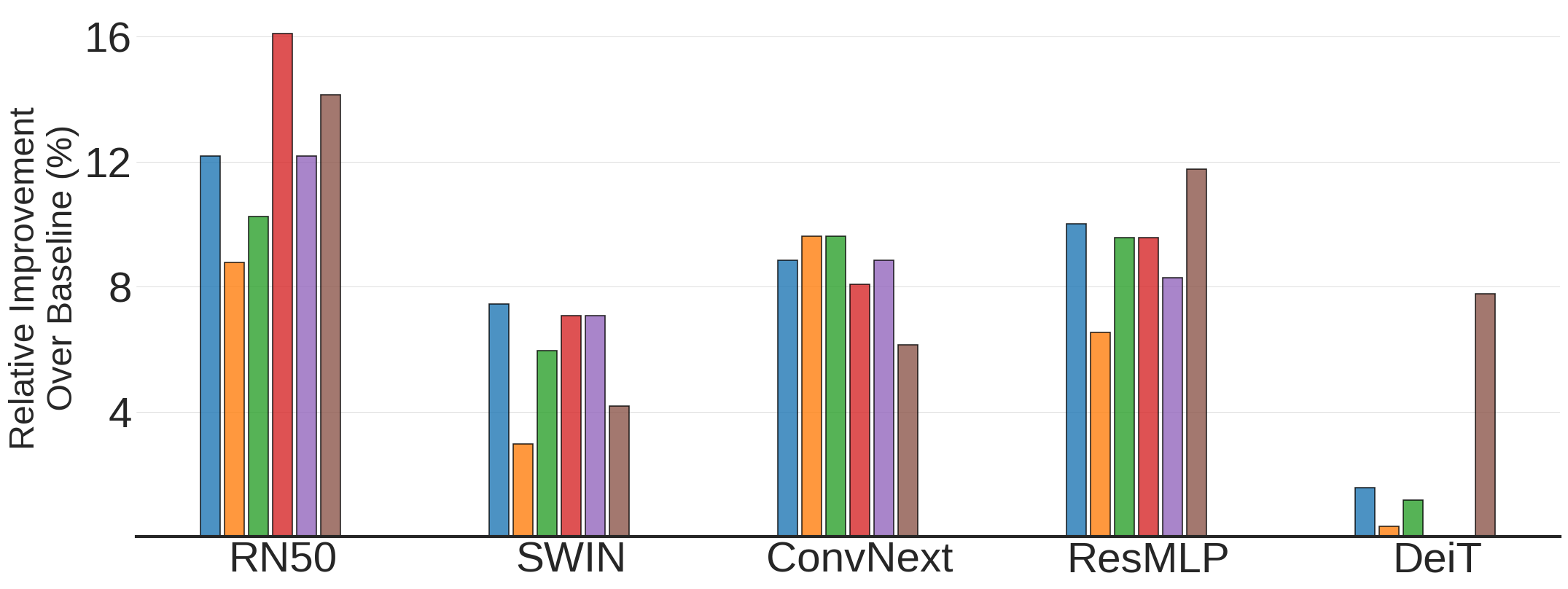}
        \vspace{-16pt}
        \captionsetup{width=\textwidth, font=footnotesize}
        \subcaption{{OfficeHome} }
        \label{fig:arch_ofh_ra}
    \end{minipage}
    \captionsetup{width=\textwidth, font=footnotesize}
    \caption{{\bf Effect of backbone.} For each of the UDA methods, we show the gain in accuracy relative to a baseline trained only using source-data for (\subref{fig:arch_dnet_cs}) DomainNet and (\subref{fig:arch_ofh_ra}) OfficeHome datasets. } 
    \label{fig:arch_da_ablation_supp}
    \vspace{-2em}
\end{center}
\end{figure*}

\begin{figure*}[!tp]
\begin{center}
    \begin{minipage}[b]{0.75\textwidth}
        \centering
        \includegraphics[width=\textwidth]{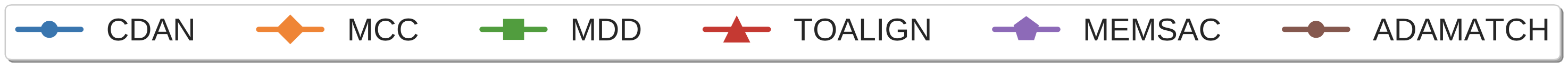}
    \end{minipage}
    
    \begin{minipage}[b]{0.46\textwidth}
        \centering
        \includegraphics[width=\textwidth]{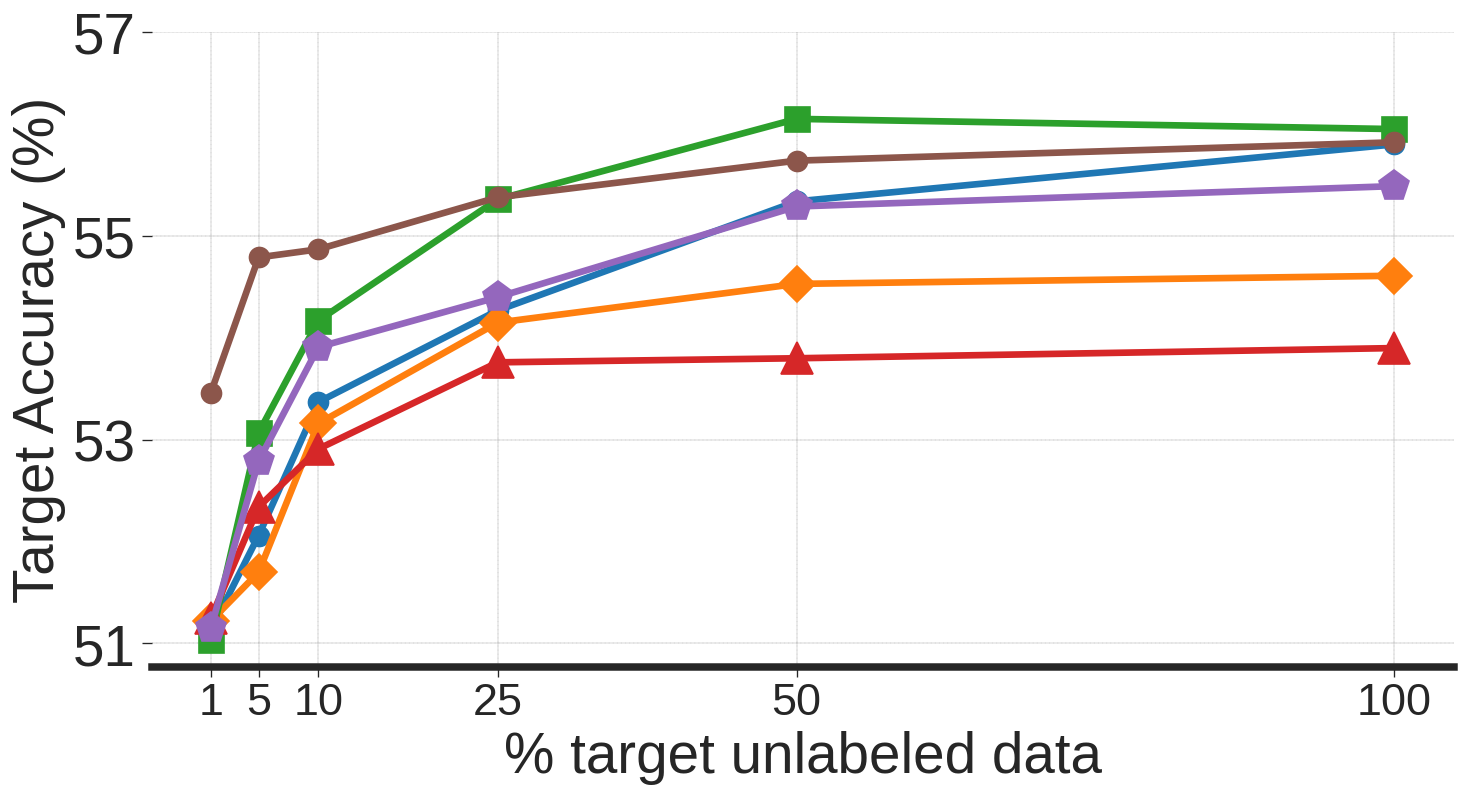}
        \captionsetup{width=\textwidth, font=footnotesize}
        \subcaption{DomainNet}
        \label{fig:datavol_rc_deit}
    \end{minipage}
    \hfill
     \begin{minipage}[b]{0.46\textwidth}
        \centering
        \includegraphics[width=\textwidth]{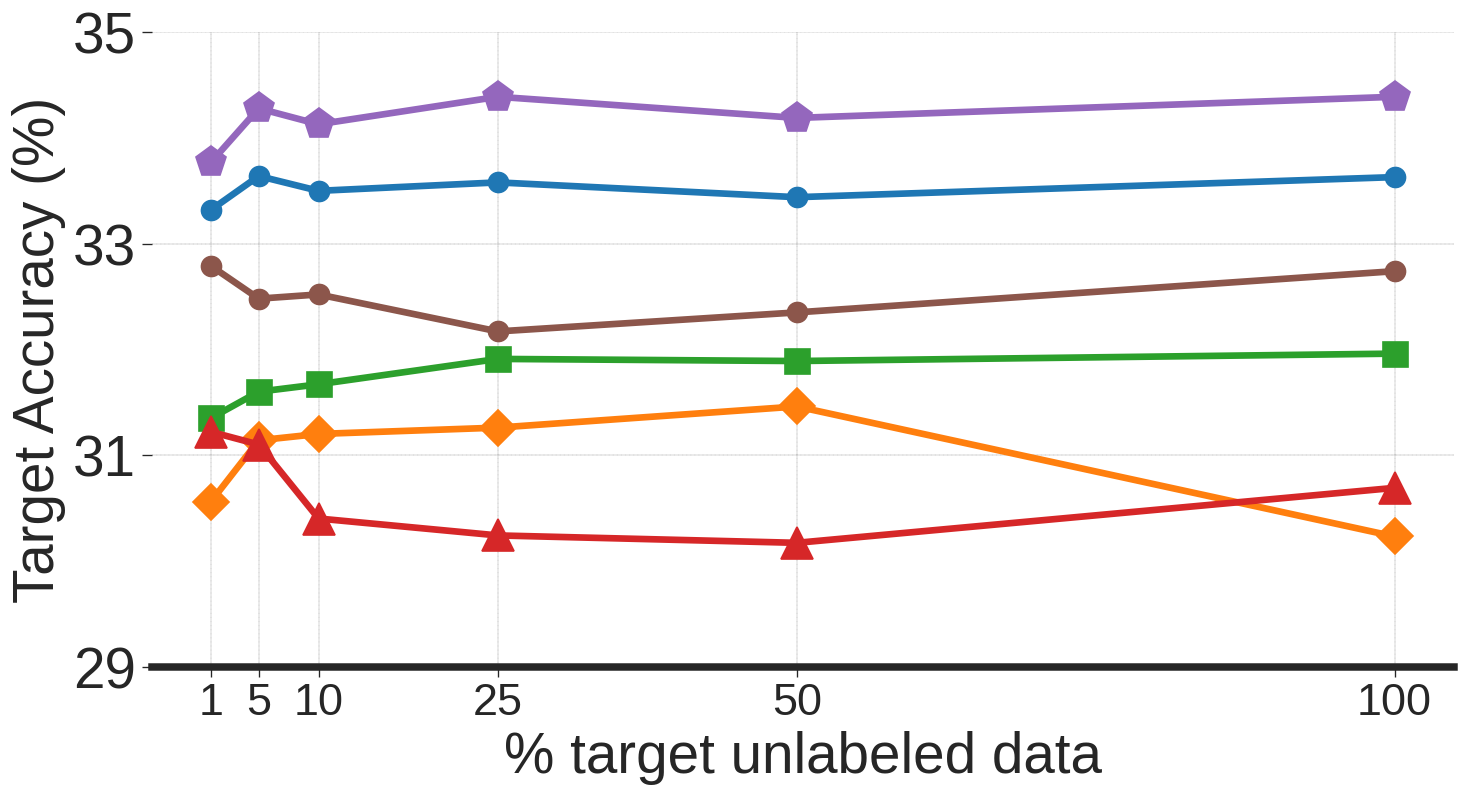}
        \captionsetup{width=\textwidth, font=footnotesize}
        \subcaption{GeoPlaces (USA$\rightarrow$Asia)}
        \label{fig:datavol_ua}
    \end{minipage}
    
    \captionsetup{width=\textwidth, font=footnotesize}
    \caption{{\bf Effect of unlabeled data } We show the effect of target unlabeled data on the target accuracy on - (\subref{fig:datavol_rc_deit}) DomainNet and (\subref{fig:datavol_ua}) GeoPlaces using a DeiT Backbone. The trends remain similar, where we observe that most UDA methods under-utilize unlabeled data.} 
    \label{fig:datavol_ablation_supp}
\end{center}
\end{figure*}

\newpara{Effect of Backbone Architecture} We examine the effect caused due to backbone on further settings from the DomainNet dataset in \cref{fig:arch_dnet_cs} and OfficeHome in \cref{fig:arch_ofh_ra}, where we show the target accuracy on each dataset.  We observe same trends as discussed in main paper with more focused transfer settings, with vision transformer architecture like Swin and Deit diminishing the benefits of most UDA methods, that otherwise yield good gains with Resnet-50 as the backbone. 

\vspace{2em}
\newpara{Amount of Unlabeled Data} As demonstrated in main paper, current UDA methods under-utilize unlabeled data, and the performance saturates even when more unlabeled data is accessible to the algorithms. We examine this trend for other settings in DomainNet dataset as well, and show the results from DomainNet in \cref{fig:datavol_ablation_supp} using a DeiT backbone. We use a DeiT backbone for this comprehensive experiment since it converges faster and hence needs lesser GPU hours during training.  
We also show the scaling trends for adaptation another completely different kind of dataset GeoPlaces~\cite{kalluri2023geonet} in \cref{fig:datavol_ua}, where we observe that unlabeled data rarely helps, even hurting the adaptation accuracy in some cases.

\begin{figure*}[t]
\begin{center}

    \begin{minipage}[b]{0.35\textwidth}
        \centering
        \includegraphics[width=\textwidth]{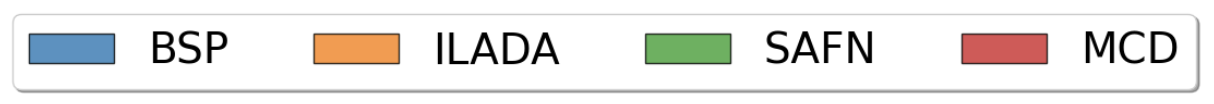}
    \end{minipage}
    
    \begin{minipage}[b]{0.24\textwidth}
        \centering
        \includegraphics[width=\textwidth]{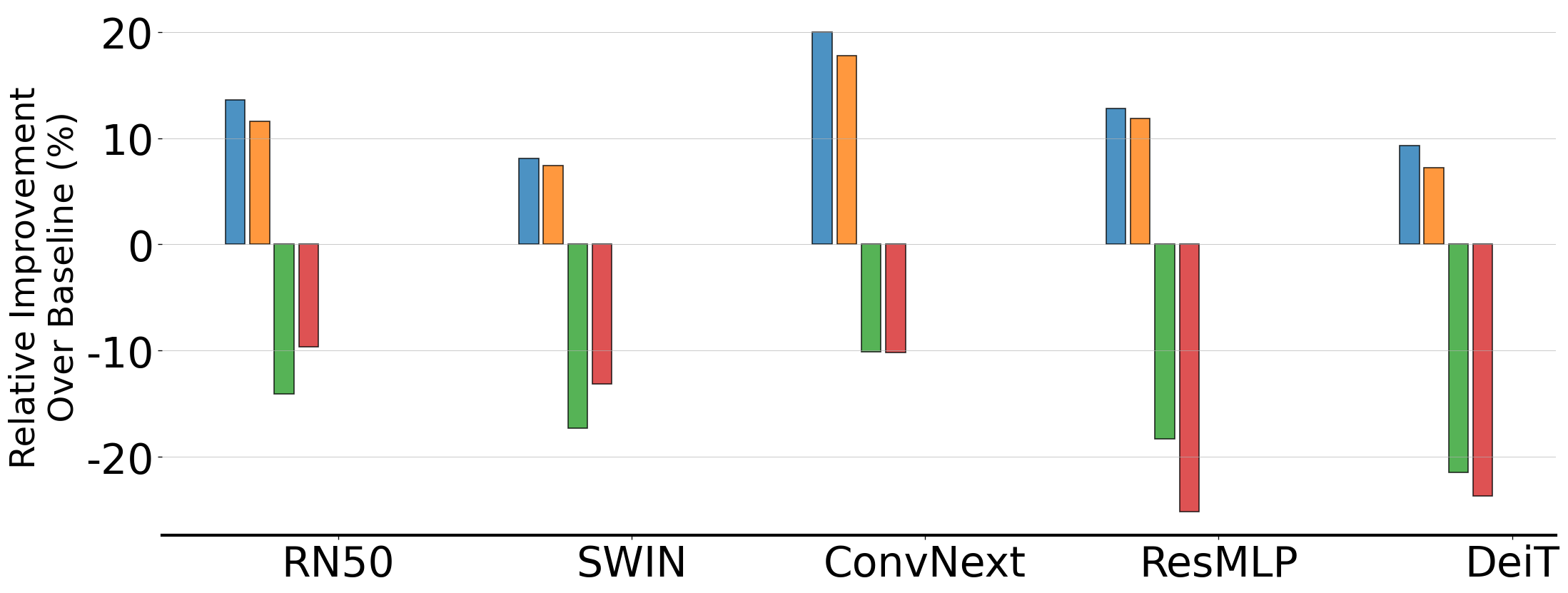}
        \vspace{-16pt}
        \captionsetup{width=\textwidth, font=footnotesize}
        \subcaption{{DomainNet} }
        \label{fig:arch_dnet_rc_x}
    \end{minipage}
    \hfill
     \begin{minipage}[b]{0.24\textwidth}
        \centering
        \includegraphics[width=\textwidth]{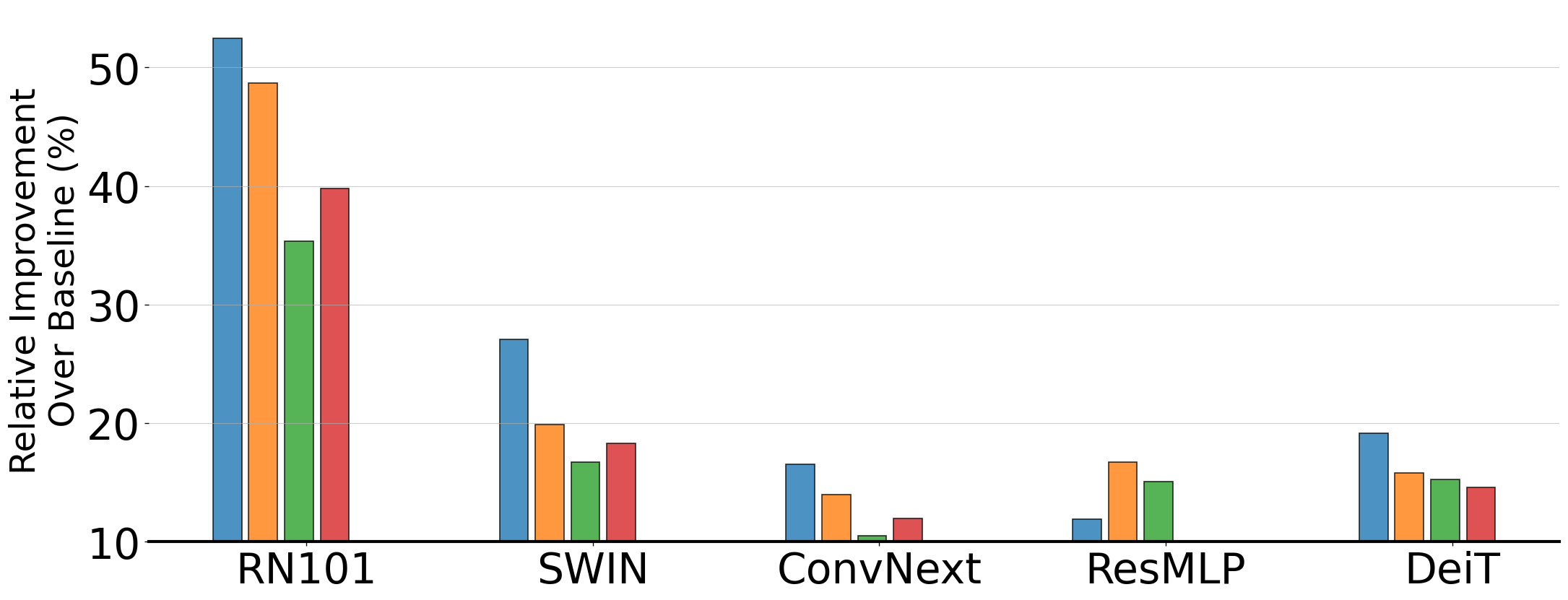}
        \vspace{-16pt}
        \captionsetup{width=\textwidth, font=footnotesize}
        \subcaption{{visDA} }
        \label{fig:arch_dnet_visda_x}
    \end{minipage}
    \hfill
    \begin{minipage}[b]{0.24\textwidth}
        \centering
        \includegraphics[width=\textwidth]{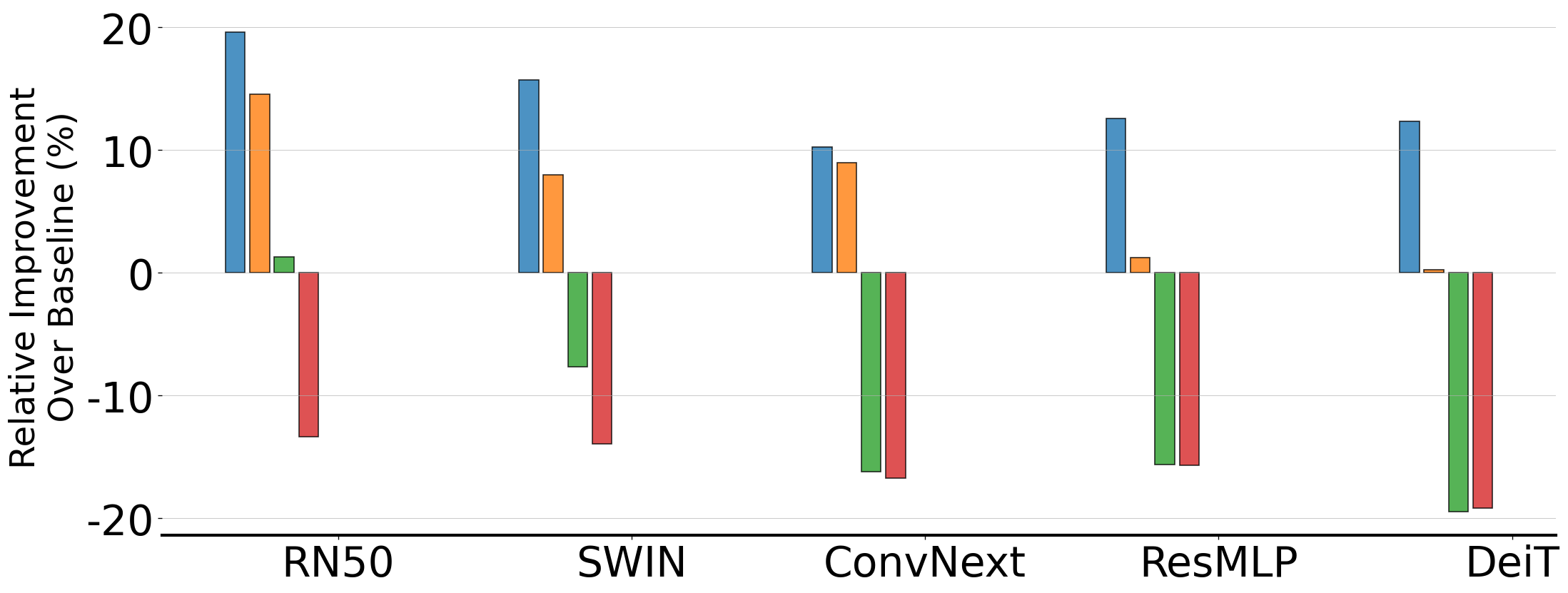}
        \vspace{-16pt}
        \captionsetup{width=\textwidth, font=footnotesize}
        \subcaption{{CUB200} }
        \label{fig:arch_cub_cd_x}
    \end{minipage}
    \hfill
    \begin{minipage}[b]{0.24\textwidth}
        \centering
        \includegraphics[width=\textwidth]{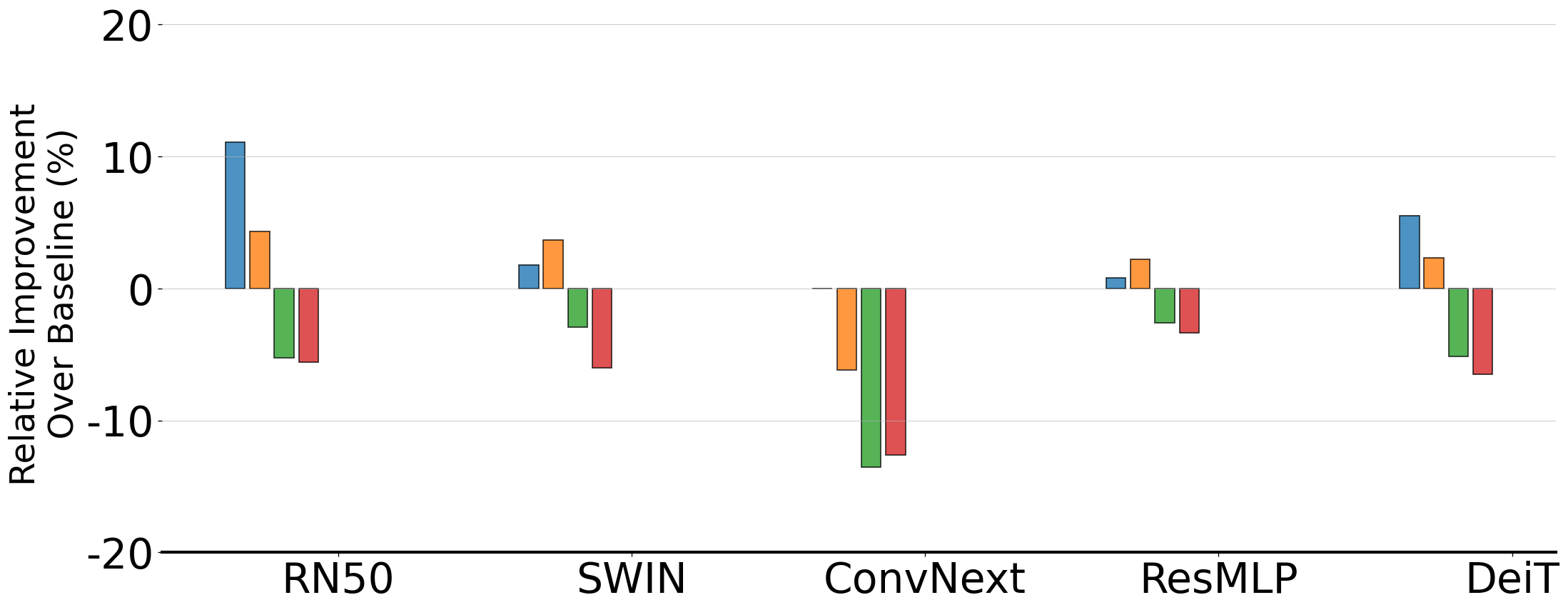}
        \vspace{-16pt}
        \captionsetup{width=\textwidth, font=footnotesize}
        \subcaption{{OfficeHome} }
        \label{fig:arch_ofh_ap_x}
    \end{minipage}
    \captionsetup{width=\textwidth, font=footnotesize}
    \caption{{\bf Newer backbones give limited returns or perform worse than baseline. } For each of the UDA methods, we show the gain in accuracy relative to a baseline trained only using source-data. For methods like SAFN~\cite{xu2019larger} and MCD~\cite{saito2018maximum}, we observe that the relative improvement over a source-only baseline is negative in most cases. Further, the gains observed by other methods like BSP~\cite{chen2019transferability} and ILADA~\cite{sharma2021instance} are not same across architectures.} 
    \label{fig:arch_da_ablation_x}
\end{center}
\end{figure*}

\begin{figure*}[t]
\begin{center}
    \begin{minipage}[b]{0.35\textwidth}
        \centering
        \includegraphics[width=\textwidth]{figures/arch_plot_legend_extra.png}
    \end{minipage}
    
    \begin{minipage}[b]{0.4\textwidth}
        \centering
        \includegraphics[width=\textwidth]{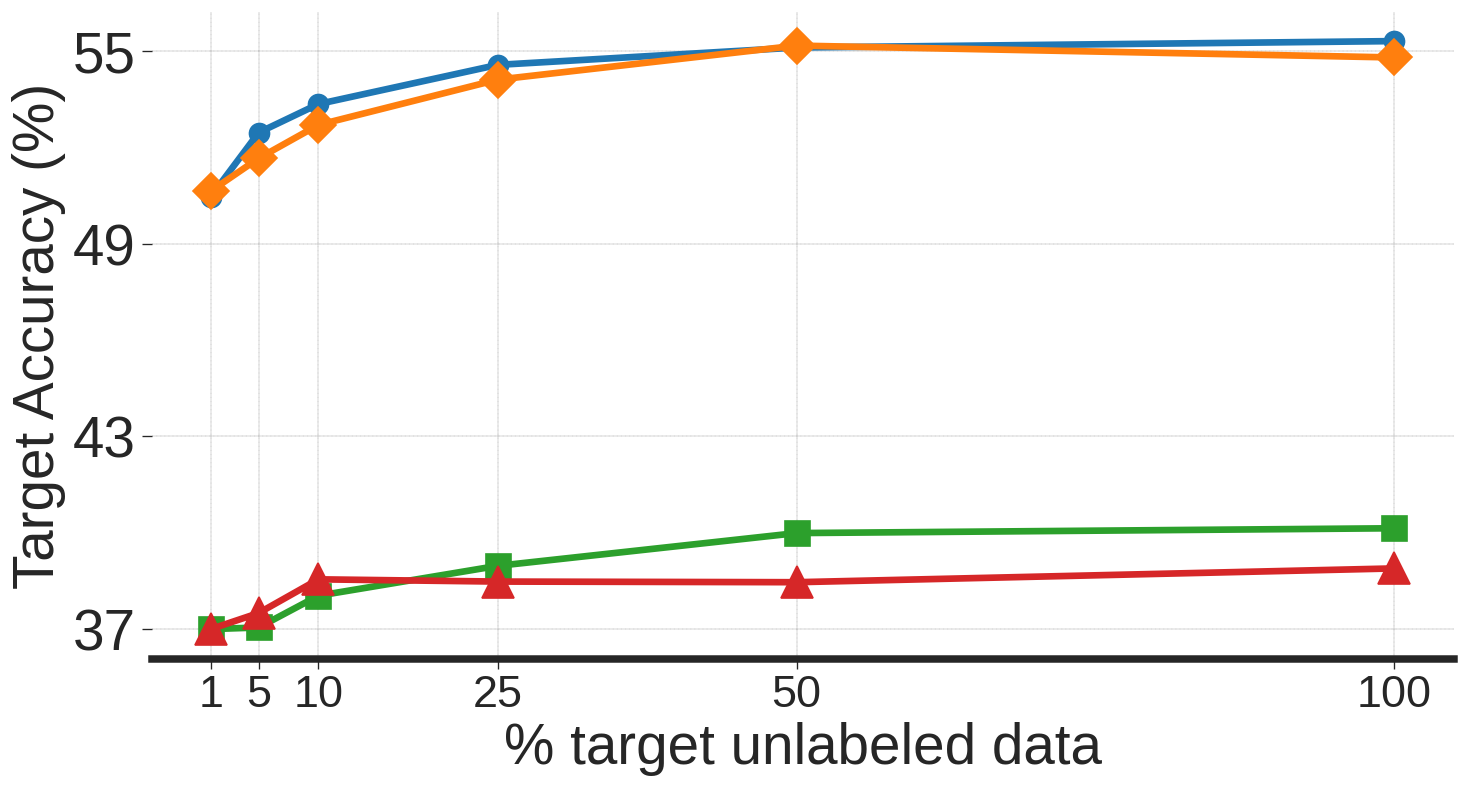}
        \captionsetup{width=\textwidth, font=footnotesize}
        \subcaption{DomainNet}
        \label{fig:datavol_rc_x}
    \end{minipage}
    ~~
    \begin{minipage}[b]{0.4\textwidth}
        \centering
        \includegraphics[width=\textwidth]{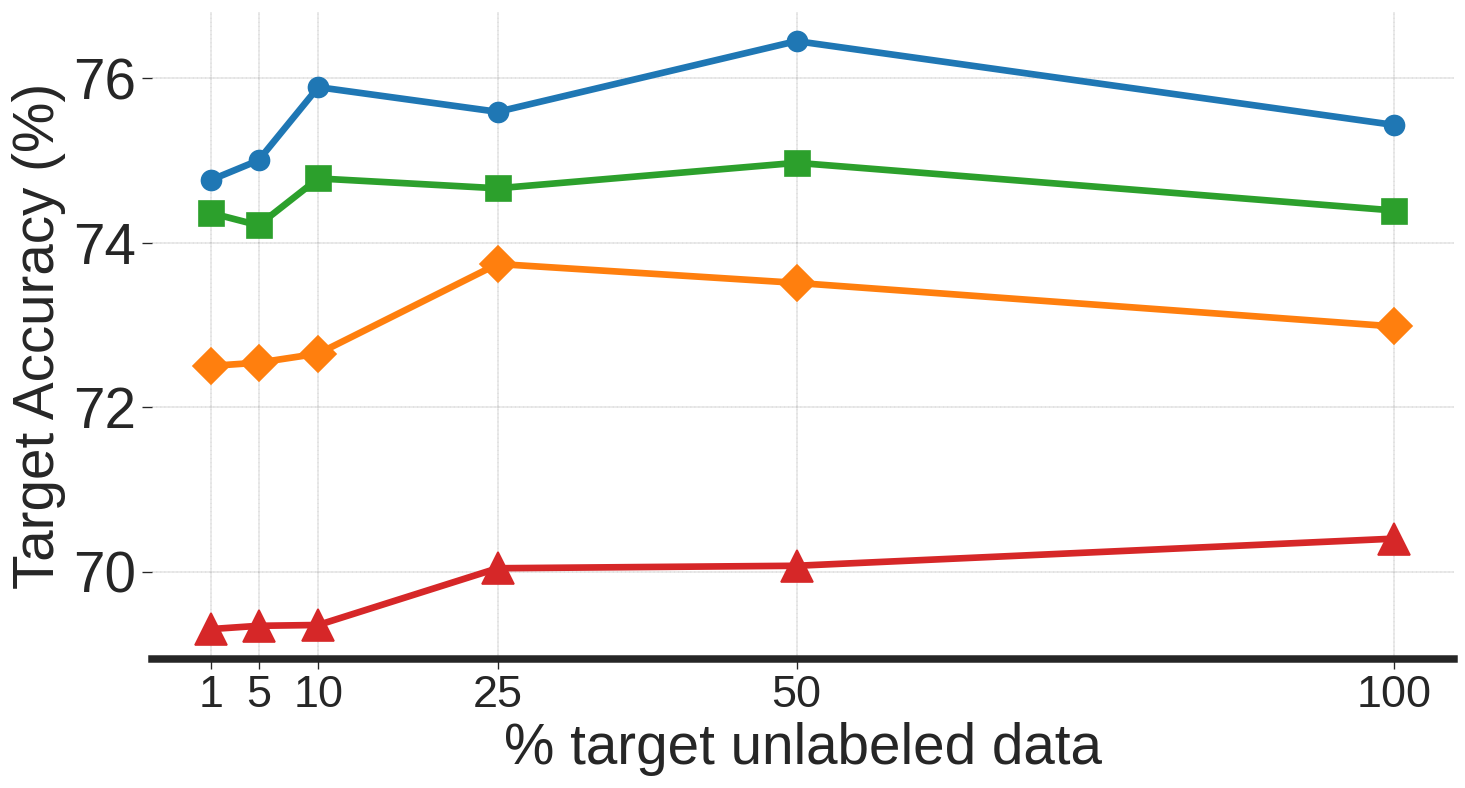}
        \captionsetup{width=\textwidth, font=footnotesize}
        \subcaption{VisDA}
        \label{fig:datavol_visda_x}
    \end{minipage}
    
    \captionsetup{width=\textwidth, font=footnotesize}
    \caption{{\bf Unlabeled Data-efficiency of UDA algorithms} Across both DomainNet and visDA datasets, the performance of UDA methods exhibits diminishing returns with increasing amounts of unlabeled data. In most cases, utilizing only 25\% of the available unlabeled data results in a performance drop of less than 1\%, suggesting that collecting additional unlabeled data is unlikely to yield significant improvements for these methods.
    } 
    \label{fig:datavol_ablation_x}
    \vspace{-2em}
\end{center}
\end{figure*}

\section{Results Using Additional UDA Methods}
\label{sec:other_results}

In addition to the wide variety of UDA methods studied in our main paper, we show results using four additional adaptation methods: BSP~\cite{chen2019transferability}, ILADA~\cite{sharma2021instance}, SAFN~\cite{xu2019larger} and MCD~\cite{saito2018maximum}. The observations for the effect of backbone architecture is presented in \cref{fig:arch_da_ablation_x} and the study for the effect of unlabeled target domain data is presented in \cref{fig:datavol_ablation_x}. 

\newpara{{\bf UDA methods are not always compatible with newer backbones.}} 
Resonating with the observations made in the main paper, we show in \cref{fig:arch_da_ablation_x} that the gains obtained by UDA method are not independent of the backbone. For instance, on CUB200 dataset, BSP~\cite{chen2019transferability} and ILADA~\cite{sharma2021instance} gives 20\% and 15\% relative gain respectively, but using DeiT diminishes these gains to 12\% and 3\% respectively. Similarly, on visDA, the improvements using ResNet is much higher than improvements offered on other backbones like ConvNext and DeiT. Moreover, as demonstrated in previous research~\cite{kalluri2022memsac}, other unsupervised domain adaptation (UDA) algorithms, such as SAFN~\cite{xu2019larger} and MCD~\cite{saito2018maximum}, under-perform compared to a source-only baseline, and the disparity worsens when employing these algorithms with newer architectures.

\newpara{Adding More Unlabeled Data is Not Beneficial for UDA} 
From \cref{fig:datavol_ablation_x}, the performance of the additional adaptation methods studied also plateaus quickly, reaching near saturation after utilizing only 20\% of the available unlabeled data. Further addition of unlabeled data yields negligible performance gains. This suggests that collecting additional unlabeled data is unlikely to yield significant improvements for these methods, corroborating the observations noted in the main paper for several other UDA methods.

\begin{figure*}[!t]
\begin{center}
    \begin{minipage}[b]{0.75\textwidth}
        \centering
        \includegraphics[width=\textwidth]{figures/datavol_legend_with_daln.png}
    \end{minipage}

    \begin{minipage}[b]{0.46\textwidth}
        \centering
        \includegraphics[width=\textwidth]{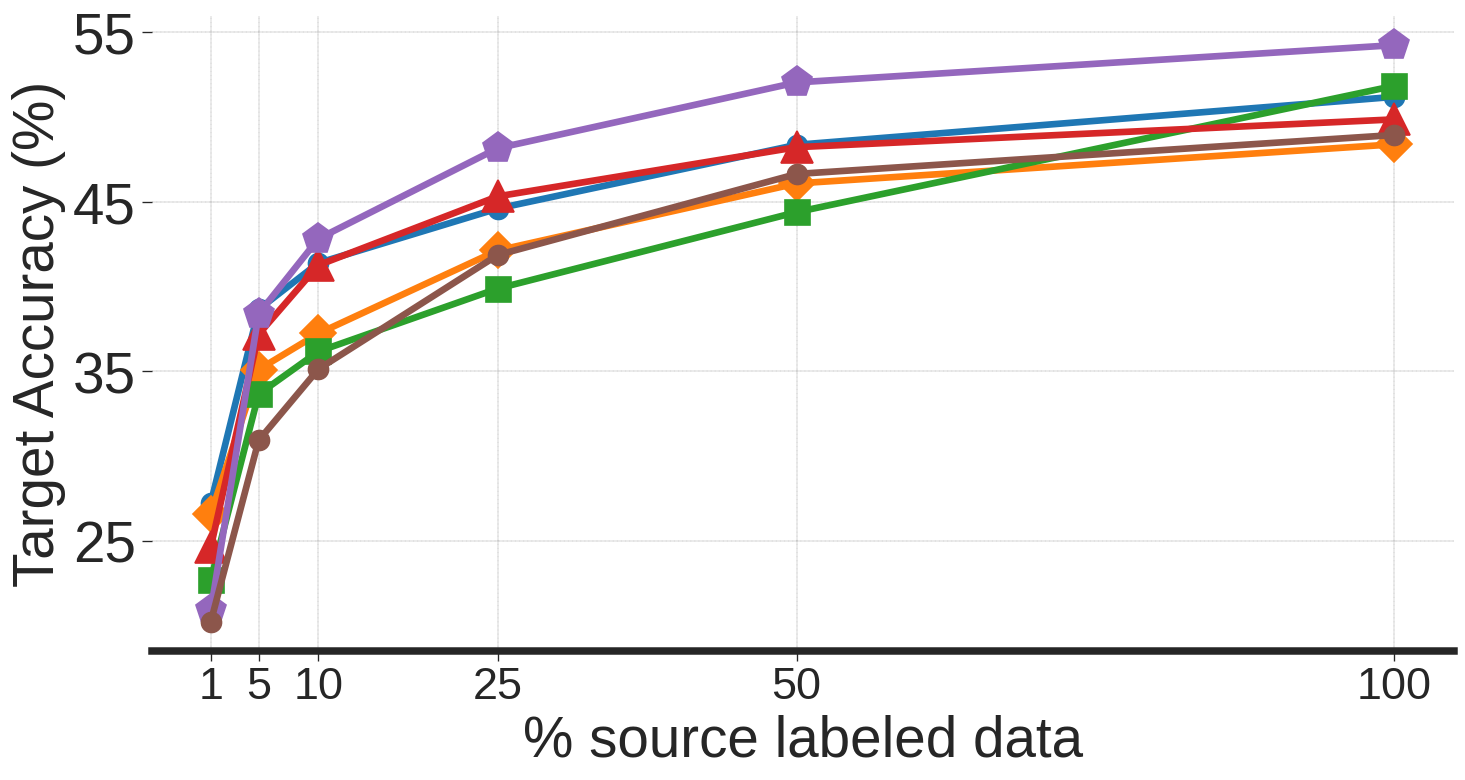}
        \captionsetup{width=\textwidth, font=footnotesize}
        \subcaption{Source Labeled Data}
        \label{fig:source_lab}
    \end{minipage}
    \hfill
    \begin{minipage}[b]{0.46\textwidth}
        \centering
        \includegraphics[width=\textwidth]{figures/datavol_rc.png}
        \captionsetup{width=\textwidth, font=footnotesize}
        \subcaption{Target Unlabeled Data}
        \label{fig:target_unlab}
    \end{minipage}     
    
    \captionsetup{width=\textwidth, font=footnotesize}
    \caption{{\bf Source labels vs. Target unsupervised data } We show that collecting more labels from source dataset, even when it is from a different domain, has a more profound influence on the target accuracy (\subref{fig:source_lab}) compared to collecting more unlabeled data from the target domain using current UDA methods (\subref{fig:target_unlab}). Results shown on Real$\rightarrow$Clipart setting from DomainNet dataset.} 
    \label{fig:datavol_ablation_source_sup}
\end{center}
\end{figure*}

\section{Source Labeled vs. Target Unlabeled Data}
In the main paper, we showed that volume of target data has minimal effect on the target accuracy after a certain point. 
To compare this with the importance held by source labels in determining the target accuracy, we conduct an experiment by using subsets of source labeled data, while using the full target unlabeled data each time. Specifically, we use $\{1,5,10,25,50,100\}\%$ of source labels and train the UDA methods on each subset. We run three random seeds and plot the mean accuracy in \cref{fig:datavol_ablation_source_sup}. We observe that the scaling trends of target accuracy with respect to source labeled data are much more favorable towards improving performance. For example, doubling the number of source labels from 50\% to 100\% improves target accuracy by $\sim 9\%$ on average across UDA methods. In contrast, the improvement in doubling the target unlabeled data from 50\% to 100\% is less than $0.5\%$ on average. This confirms the fact that labels have a more pronounced impact on target accuracy even when they arise from a different domain, compared to unlabeled data from the same domain. 

\section{Results using TinyImageNet}

\begin{figure*}[!t]
\begin{center}

    \begin{minipage}[b]{0.54\textwidth}
        \centering
        \includegraphics[width=\textwidth]{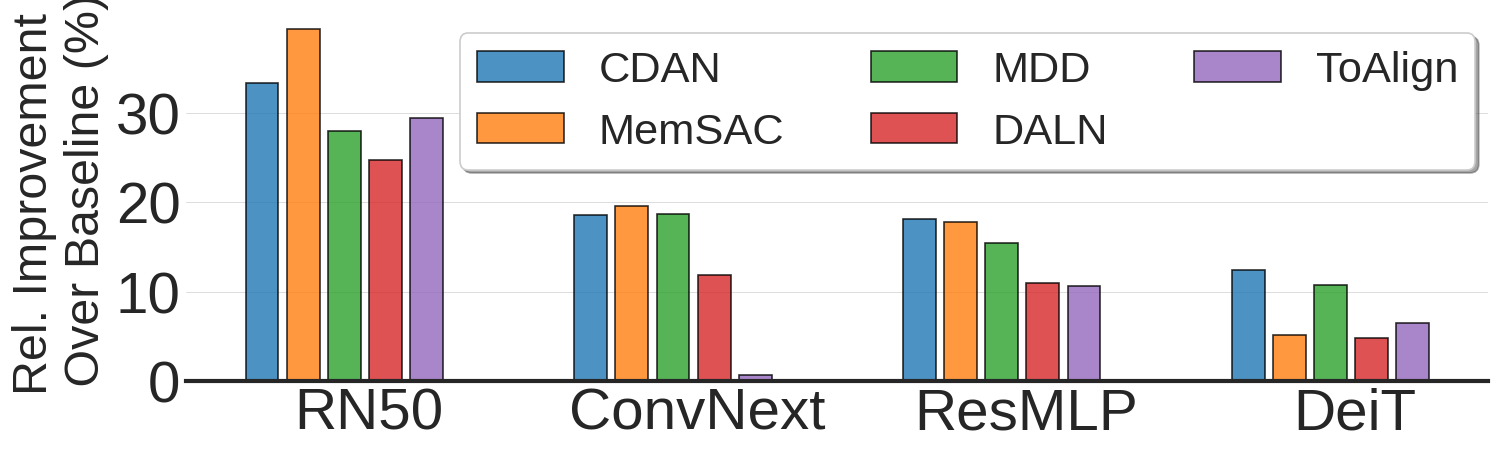}
        \captionsetup{width=\textwidth, font=footnotesize}
        \subcaption{Varying backbones}
        \label{fig:imnet_arch}
    \end{minipage}
    \hfill
    \begin{minipage}[b]{0.4\textwidth}
        \centering
        \includegraphics[width=\textwidth]{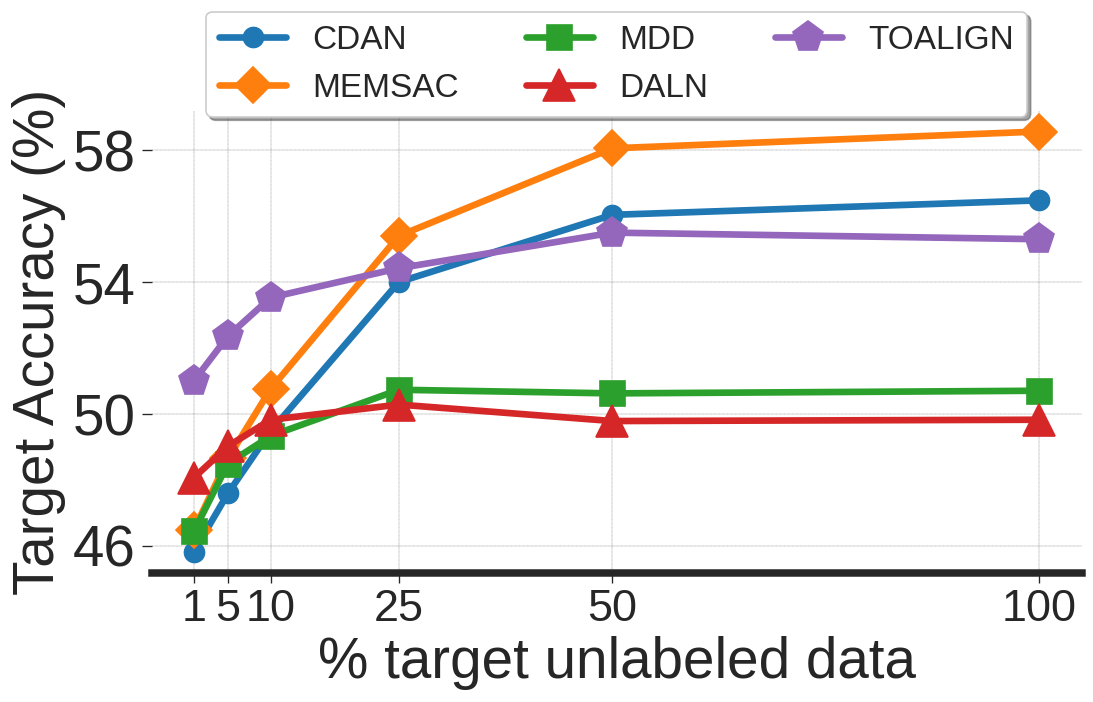}
        \captionsetup{width=\textwidth, font=footnotesize}
        \subcaption{Varying target unlabeled data}
        \label{fig:imnet_vol}
    \end{minipage}     
    
    \captionsetup{width=\textwidth, font=footnotesize}
    \caption{{\bf Results on TinyImageNet vs. TinyImageNet-C } We show the similar observations regarding backbone architectures and data volume hold also for a non-standard adaptation dataset. We use images from TinyImageNet as the source and \textit{snow-3} perturbations from TinyImageNet-C as the target.} 
    \label{fig:tinyimagenet}
\end{center}
\end{figure*}

To further examine the presented trends on non-standard adaptation datasets, we show results using images from the TinyImageNet dataset as the source domain and \textit{snow} perturbations from TinyImageNet-C~\cite{hendrycks2019benchmarking} as the target domain. We train models using the 200 classes in each dataset, and use report accuracy on the target domain. In \cref{fig:tinyimagenet}, we show that the broad trends observed for other adaptation datasets also hold for this novel setting. Specifically, from \subref{fig:imnet_arch}, adaptation gains are much lesser with recent architectures (like \textit{ConvNext} and \textit{DeiT}) and from \subref{fig:imnet_vol}, performance saturates in-spite of adding unlabeled data, further corroborating the main inferences from our study.

\section{Training Details}
\label{sec:training_details}

\newpara{Architecture-specific training details} In our ablation on benchmarking UDA across architectures, we use all pre-trained checkpoints from the timm library, and all of them are pre-trained on ImageNet-1k. Across the architectures, we uniformly use a batch size of 32, SGD optimizer with an initial learning rate of 0.003 and cosine decay. It might be possible that ViT models benefit from other algorithms such as Adam~\cite{zhang2020adaptive}, which we do not explore in this paper. For data augmentation, we first resize the images so that the shorter size is 256 and then choose a random $224 \times 224$ crop followed by random horizontal flip. However, we use a crop size of 256 instead of 224 for Swin transformer due to its input size. We train the networks for a total of 75k iterations on DomainNet and CUB200 with validation performed at every 5k steps, and for 30k iterations on the smaller OfficeHome dataset with validation at every 500 steps. We use early stopping on the test set to choose the best accuracy. 

For the classifier, we use a 2-layer MLP with a hidden dimension of 256. The input dimension for the MLP, though, varies depending on the output dimension of the backbone architecture used. For Resnet-50, it is 2048, for Swin-t and ConvNext-t it is 768 and for Deit-s and ResMLP-s it is 384. 

\subsection{Unsupervised Pre-Training Network Details}

We use the official repositories for SwAV, MoCo-v3, MAE to pre-train the models on our datasets. Note that we subsample an image set of 1M images from ImageNet, Places205 and iNat2021 to normalize the effects of data volume, using a per-class sampling strategy. We use the official repositories for Swav, MoCo-V3 and MAE, and use the code for supervised pre-training from \href{https://github.com/pytorch/examples/tree/main/imagenet}{PyTorch}. We train Swav for 150 epochs, MoCo-v3 for 250 epochs, MAE for 400 epochs and supervised pre-training for 90 epochs. The training for all the methods is performed on 8 GPUs with a total batch size of 1024 in each case. For all other hyperparameters, we follow the ones recommended in the respective repositories. 

\end{document}